\documentclass[10pt,twocolumn,letterpaper]{article}

\usepackage{3dv}
\makeatletter
\@namedef{ver@everyshi.sty}{}
\makeatother
\usepackage{tikz}
\usepackage{times}
\usepackage{epsfig}
\usepackage{graphicx}
\usepackage{amsmath}
\usepackage{amssymb}
\usepackage{times}
\usepackage{epsfig}
\usepackage{graphicx}
\usepackage{amsmath}
\usepackage{amssymb}
\usepackage{bbding}
\usepackage{multirow}
\usepackage{subcaption}
\usepackage{gensymb}
\usepackage{tikz}
\usepackage{comment}
\usepackage{pifont}
\usepackage{tabularx}
\usepackage{hyperref}
\usepackage{color}
\usepackage{url}

\threedvfinalcopy 

\def\threedvPaperID{****} 

\ifthreedvfinal\fi
\begin{document}

\newcolumntype{C}[1]{>{\centering}m{#1}}
\title{A Flexible Multi-view Multi-modal Imaging System for Outdoor Scenes}

\author{Meng Zhang
\qquad Wenxuan Guo
\qquad Bohao Fan
\qquad Yifan Chen\\
\qquad Jianjiang Feng
\qquad Jie Zhou\\
Department of Automation, BNRist, 
Tsinghua University, Beijing, China\\
{\tt\small \{zhangm20,gwx22,fbh19,chenyf21\}@mails.tsinghua.edu.cn \qquad \{jfeng,jzhou\}@tsinghua.edu.cn}}

\maketitle
\thispagestyle{empty}

\begin{abstract}
Multi-view imaging systems enable uniform coverage of 3D space and reduce the impact of occlusion, which is beneficial for 3D object detection and tracking accuracy. However, existing imaging systems built with multi-view cameras or depth sensors are limited by the small applicable scene and complicated composition. In this paper, we propose a wireless multi-view multi-modal 3D imaging system generally applicable to large outdoor scenes, which consists of a master node and several slave nodes. Multiple spatially distributed slave nodes equipped with cameras and LiDARs are connected to form a wireless sensor network. While providing flexibility and scalability, the system applies automatic spatio-temporal calibration techniques to obtain accurate 3D multi-view multi-modal data. This system is the first imaging system that integrates mutli-view RGB cameras and LiDARs in large outdoor scenes among existing 3D imaging systems. We perform point clouds based 3D object detection and long-term tracking using the 3D imaging dataset collected by this system. The experimental results show that multi-view point clouds greatly improve 3D object detection and tracking accuracy regardless of complex and various outdoor environments.
\end{abstract}

\section{Introduction}
A 3D dynamic imaging system is fundamental to observe, understand and interact with the world. In recent years, a large number of work~\cite{DBLP:journals/pami/salsa,DBLP:journals/access/human4d,DBLP:journals/pami/human36m,DBLP:conf/cvpr/JooPS14,DBLP:journals/pami/cmu,DBLP:conf/wacv/mhad,DBLP:journals/ijcv/SigalBB10} has been devoted to indoor 3D imaging systems, and the data obtained by these systems have made great contributions to the research work on accurate human pose estimation and mesh generation. If dynamic 3D imaging data of large-scale outdoor scenes can be obtained, it will be extremely beneficial to the application scenes such as security surveillance, sporting game analysis, and cooperative vehicle-infrastructure system (CVIS). 
\begin{figure}[h]
    \centering
    \includegraphics[width=.9\linewidth]{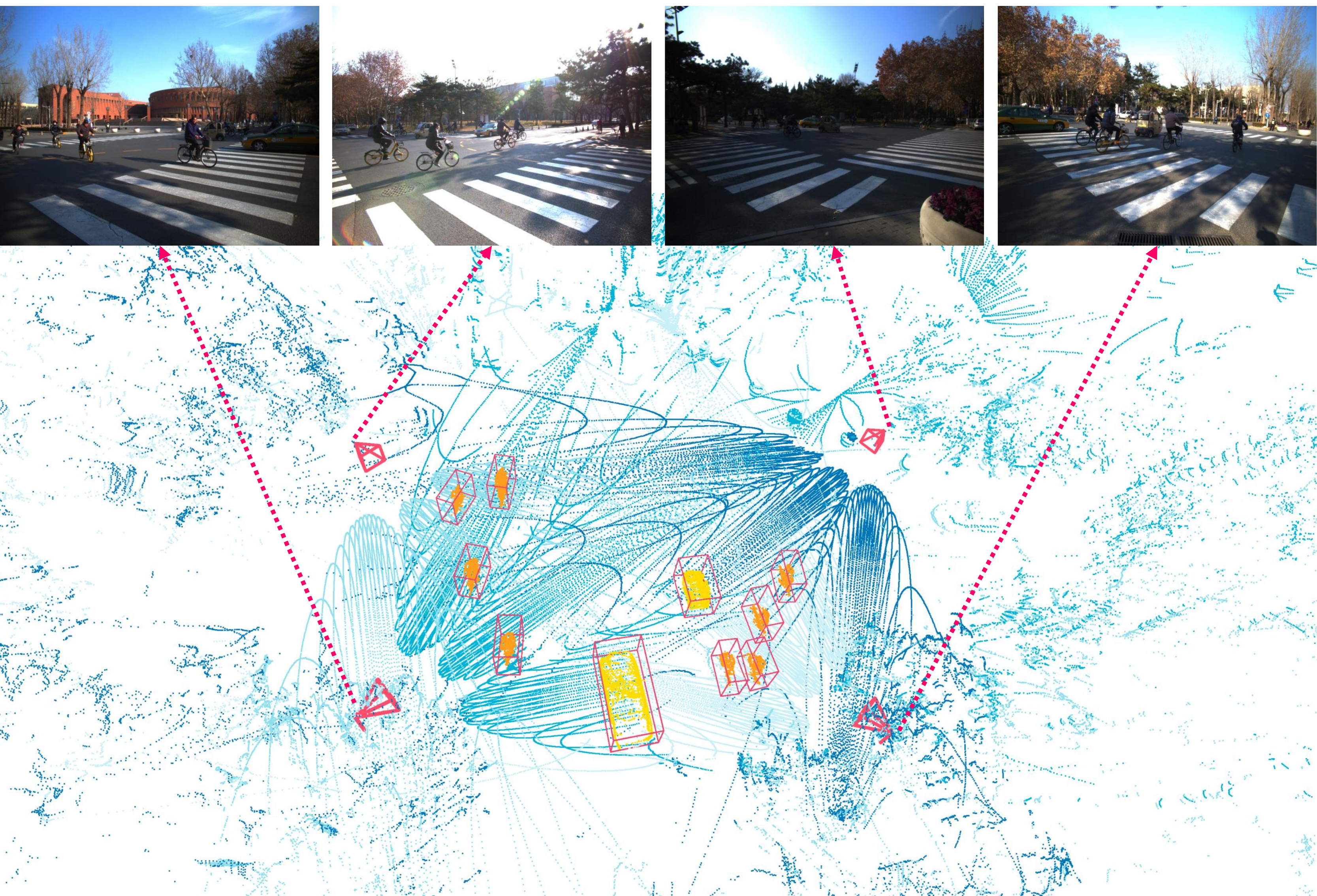}
    \caption{3D imaging data collected by our system at a crossroad. The multi-view point cloud, which is colored different shades of blue, is from four different LiDARs. The points of pedestrians are colored orange and points of cars are colored yellow. Four slave nodes are presented by the red rectangular pyramids, RGB images of which are displayed at the top.}
    \label{fig:intro}
    
\vspace{-.3cm}
\end{figure}

However, existing indoor 3D imaging systems are greatly inapplicable to outdoor large-scale scenes. First, most of the existing 3D imaging systems apply epipolar geometry estimation of multiple RGB cameras or utilize depth cameras to obtain depth data. The depth provided by epipolar geometry leads to a great absolute error from a distance, while depth sensors like Microsoft Kinect and Intel Realsense only provide a measurement range of about five meters. In addition, RGB cameras and depth sensors are apt to be disturbed by lighting conditions and adverse weather in complex and varied outdoor scenes. Apart from the limitations of RGB cameras and depth sensors, existing 3D imaging systems are restricted by the tangled and fixed system architecture, as well as the strict electric and network conditions. Most of the sensors are either permanently attached to the scene's walls and roof, or to a difficult-to-move assembly structure. In an ideal indoor environment, existing systems can easily obtain high-quality 3D imaging data by utilizing sufficient lighting conditions and a large number of cables to connect and control the sensors. Unfortunately, the vast majority of large outdoor scenes are difficult to meet similar conditions. Common outdoor scenes, such as squares and streets, have a scene size of more than 50 meters. Traditional 3D imaging systems are 
hard to deploy in outdoor scenes because the ground, wall, electric, and network conditions vary widely. Furthermore, collecting sufficient data in varied outdoor scenes calls for a system that is easy to disassemble and set up.

In this paper, we propose a portable wireless multi-view multi-modal 3D imaging system against the technical challenges in large outdoor scenes. We introduce affordable LiDAR sensors to build a multi-modal sensor network that can be used in challenging environments. The point cloud data, which can be obtained at a distance of 200 meters, is combined with the rich texture information obtained by the camera as an additional feature.
Aiming at the applications in complex outdoor scenes, our system is composed of movable slave nodes that are easy to assemble and disassemble. Furthermore, the 3D imaging system can freely expand or squeeze the node capacity due to the distributed multi-node network. We utilize a wireless network to combine each node for flexibility and design a GPS-based multi-node time synchronization system to overcome the time error in the wireless system. To avoid manual calibration among nodes in a new scene, we propose an automatic spatial calibration paradigm based on point cloud registration. 

Finally, the system is put to the test in several different outdoor scenes to collect 3D imaging data. As shown in Fig.~\ref{fig:intro}, we built a system including four slave nodes to collect multi-view point cloud data and images, which were then annotated with 3D bounding boxes and track ids. On the collected dataset, we evaluated 3D object detection and multi-object tracking and came up with a promising result. The hardware design, the code and the collected 3D imaging data, which includes annotated spatio-temporally aligned point cloud and images, are publicly available online at:  \href{http://ivg.au.tsinghua.edu.cn/dataset/THU-MultiLiCa/THU-MultiLiCa.html}{\url{http://ivg.au.tsinghua.edu.cn/dataset/THU-MultiLiCa/THU-MultiLiCa.html}}.

\section{Related Work}
The 3D imaging and understanding of the scene have received increasing attention in the past ten years. A brief overview of some typical systems follows.


Existing 3D imaging systems mostly relied on cameras in a small indoor scene~\cite{DBLP:journals/access/human4d,DBLP:journals/pami/human36m,DBLP:journals/pami/cmu,DBLP:conf/wacv/mhad,DBLP:journals/ijcv/humaneva}. Ionescu et al.~\cite{DBLP:journals/pami/human36m} employed multi-view RGB cameras and a Time-of-Flight (ToF) depth sensor to collect and publish the Human3.6M dataset, which contains a large set of 3D human poses in common activities. However, there is only one low-resolution depth sensor in the system and the dataset only focused on the single-person situation. Joo et al.~\cite{DBLP:journals/pami/cmu} proposed a modularized massively multiview capture system with a large number of cameras and depth sensors, named Panoptic Studio, and performed human skeletal pose estimation and motion tracking. This system utilized $480$ VGA cameras, $31$ HD cameras, and $10$ Microsoft Kinect depth sensors to form a sensor network in a structure with a radius of $5.49$ meters and a total height of $4.15$ meters. In an ideal construction environment, all sensors of Panoptic Studio are connected via a wired network and synchronously triggered by the master clock. 
However, massive sensors and complex systems also lead to the difficulty of reproduction, construction and free movement. Chatzitofis et al.~\cite{DBLP:journals/access/human4d} collected the Human4D dataset, consisting of $50,000$ multi-view 3D imaging data in a room equipped with 24 cameras rigidly placed on the walls and 4 Intel RealSense D415 depth sensors. In this capturing system, four depth sensors were synchronized using the HW-Synced method in order to conduct related research on multi-view 3D data with high-precision spatio-temporal alignment. All above 3D imaging systems have the problems of (1) tangled and fixed system construction, (2) the limitations of RGB cameras, and (3) scene size limitation. In contrast, the 3D imaging system proposed in this paper is flexible for construction and relies on LiDARs for the accurate depth data at a distance.

Only a few dynamic 3D multi-modal imaging systems are designed for outdoor environments. Strecha et al.~\cite{DBLP:conf/cvpr/StrechaHGFT08} set up 25 high-resolution cameras outdoors for dense data acquisition and reconstruction of static outdoor scenes like palaces. The ground truth for the reconstruction was the point cloud data obtained by a LiDAR, which needs scanning for a long time and cannot handle dynamic objects. Several research groups \cite{DBLP:conf/cvpr/Wildtrack,pets,DBLP:conf/cvpr/campus} employed multi-camera system for multi-object tracking in outdoor scenes. Chavdarova et al.~\cite{DBLP:conf/cvpr/Wildtrack} collected and published Wildtrack dataset using 4 GoPro3 cameras and 3 GoPro4 cameras in a square with dense pedestrian tracking annotations. In spite of the flexibility of system, the precision of synchronization hampered the data acquisition that seven sequences was obtained with $50$ ms accuracy. Additionally, Wildtrack dataset used $1398$ manual annotated points to calibrate multiple cameras at a high labor cost, unsuitable for deployment at many different scenes. Kim et al.~\cite{DBLP:journals/ral/pedx} present a dataset titled PedX consisting of stereo images and LiDAR data of pedestrians at a complex urban crossroad. Despite the employment of stereo LiDARs on top of a car, PedX only captured point cloud from a single direction, which means it can not get the complete shape of targets and is affected by occlusion. Unlike other outdoor 3D imaging systems, our system employs multiple spatially distributed LiDARs to obtain 3D point clouds at a long distance, ensuring that 3D data is complete and accurate. The combination of point cloud and RGB images provides richer features for 3D visual analysis of dynamics in large outdoor scenes. Finally, our automatic spatio-temporal calibration method produces high-precision multi-modal 3D data while reducing labor costs significantly, which is crucial for collecting data at various sites.

\section{Proposed Imaging System}
We present a dynamic 3D imaging system to capture multi-view, multi-modal data in large outdoor scenes such as crossroads and squares. The construction of a centralized system connected by a wired network is difficult outdoors due to the electric and network conditions. In this regard, the 3D imaging system proposed in this paper is designed in a combination of centralized and distributed structures based on a wireless network for high flexibility. Further, we apply a spatio-temporal calibration method to handle the synchronization problem for the wireless system. At each modularized slave node, an RGB camera and a Livox LiDAR provide multi-modal 3D data to scan the large scene from its location. The system captures a dynamic 3D multi-view point cloud at 10 FPS with high-precision spatio-temporal alignment so that the point cloud of the object is completed from all directions to avoid occlusion as much as possible. In this section, we introduce the structure design of the system, calibration techniques, and nodes in details. More technical details about the proposed system are available in the supplementary materials.

\subsection{Structural Design}
\begin{figure}[]
    \centering
    \includegraphics[width=1\linewidth]{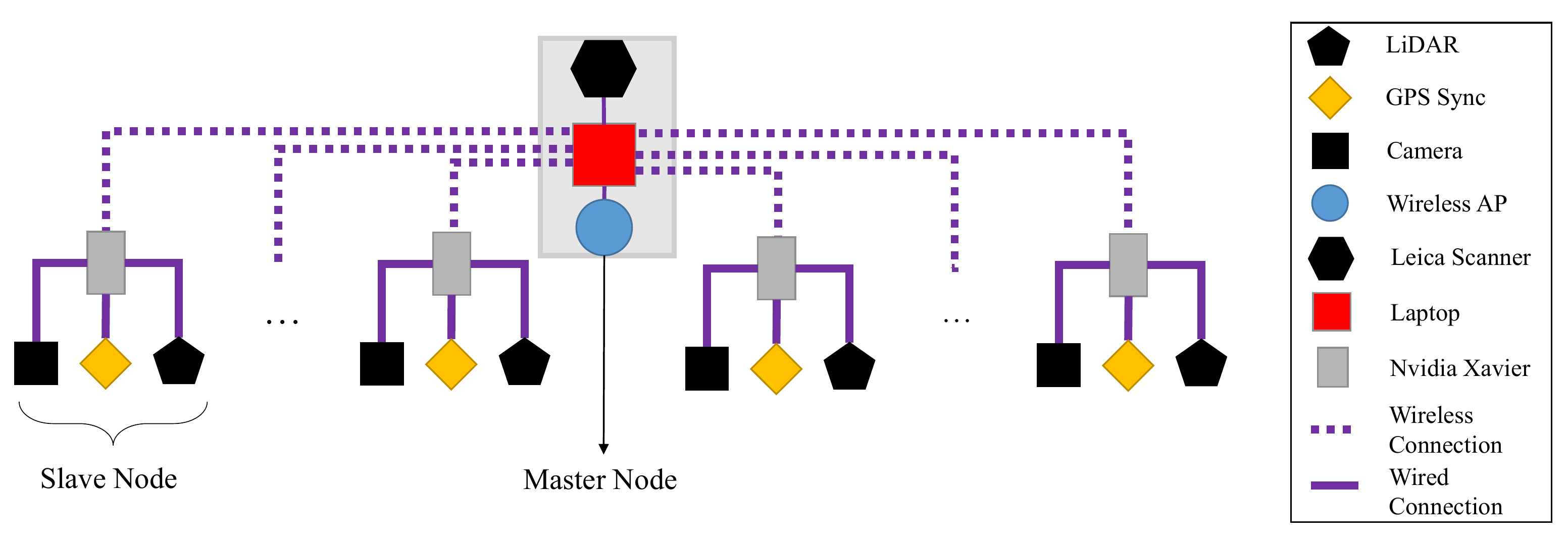}
    \caption{The system architecture of our system. Master node controls the entire system via a wireless network. Each modularized slave node utilizes an RGB camera and a Livox LiDAR to capture 3D imaging data.}
    \label{fig:structure}
\end{figure}

Our system is constructed in a centralized and distributed form~\cite{DBLP:conf/sitis/lowcostsystem}, which is made up of a master node and several slave nodes. The master node is connected to all slave nodes in order to monitor and maintain the entire system. A 3D data stream is captured and processed by each slave node independently of the others. The architecture of our system is depicted in Fig.~\ref{fig:structure}.

To manage node communication in a vast outdoor environment, we employed an wireless access point which can cover a distance of 100 meters while maintaining good signal quality and ensuring system connection stability.
The master node is a Linux laptop connected to the outdoor wireless access point via a network cable, while slave nodes, which serve as acquisition modules, are wirelessly connected to the access point. All of the nodes send image data and control signals via robot operating system (ROS). Despite a stable system connection, the communication bandwidth is insufficient to handle the enormous data stream of point cloud and image between the master node and slave nodes. As a result, slave nodes preprocess and send compressed data to the master node, while all raw point cloud data and images are stored locally.

\subsection{Spatio-temporal Calibration}\label{sec:calib}
In 3D imaging acquisition, time synchronization is crucial. Due to the outdoor environment, hardware triggering between sensors cannot be done directly using a wired connection. As for the wireless network, time synchronization is challenged by the random transmission delays. To address this problem, we present a synchronization method via software triggering and Pulse-Per-Second (PPS) signals from a global positioning system (GPS). The proposed method synchronizes the node’s system clock using the accurate time source from a low-cost GPS module. Accurate PPS signals and National Marine Electronics Association (NMEA) sentences generated by GPS modules are utilized to get a very accurate timestamp. The timestamp error between two separate GPS modules is estimated to less than one microsecond~\cite{timesync}. To align the start frames of multiple nodes, each slave node initiates the ready state but does not start acquisition after the system is up and running. The master node broadcasts a trigger signal through the wireless network to wake up the slave nodes. Then each slave node waits for the first PPS signal from GPS to acquire the first frame of images. The same PPS signal aligns the start time of 3D imaging data with a one-microsecond error. 

Traditional multi-camera calibration methods typically require specific calibration geometry or texture patterns to mark corresponding points across multiple views~\cite{DBLP:conf/cvpr/calibfusion,DBLP:journals/trob/cameracalib}, which are difficult to apply in large scenes and result in inaccuracy in spatial calibration across cameras. The spatial calibration must be performed anew if the multi-camera system is relocated. To handle the tedious and repetitive spatial calibration, we introduce a calibration paradigm based on the synchronized 3D multi-modal data. We used a Leica BLK360 scanner to collect a static dense point cloud of the scene as the world coordinate system. Then, the point cloud stream from each acquisition node is merged over ten seconds to create a dense point cloud of the scene for each slave node. We extract the FPFH feature~\cite{DBLP:conf/icra/FPFH} of different scales from the point cloud to apply a hierarchical ICP~\cite{DBLP:journals/pami/icp} method for registration. The dense point cloud of the scene from each view is automatically aligned to the Leica point cloud so as to get the transformation matrices. We calculate the intrinsic and extrinsic parameters of cameras using the Pinhole camera model~\cite{DBLP:conf/avss/intrinsic}. As the camera and LiDAR are fixed in place on each slave node, the cameras’ extrinsic parameters are only calibrated once at the beginning. Even after switching to a new scene, all transformation matrices, intrinsic and extrinsic parameters can be obtained automatically. By combining LiDAR-LiDAR and LiDAR-camera calibration, our system can complete the spatial calibration of cameras and LiDARs automatically.

\subsection{Portable Node}
The master node of our system can range from a portable laptop to a high-performance workstation depending on the tasks, and only needs to be wired to the wireless access point through a gigabit ethernet cable. All activities, such as starting and stopping acquisition of a node 50 meters away, can be immediately finished with the master node. In addition to controlling the system, the master node displays data for acquisition status monitoring. The sampled compressed images from each slave node are sent to the master node via a wireless network to show the current scene.

\begin{figure}
    \centering
    \begin{subfigure}{.75\linewidth}
    \includegraphics[width=1\linewidth]{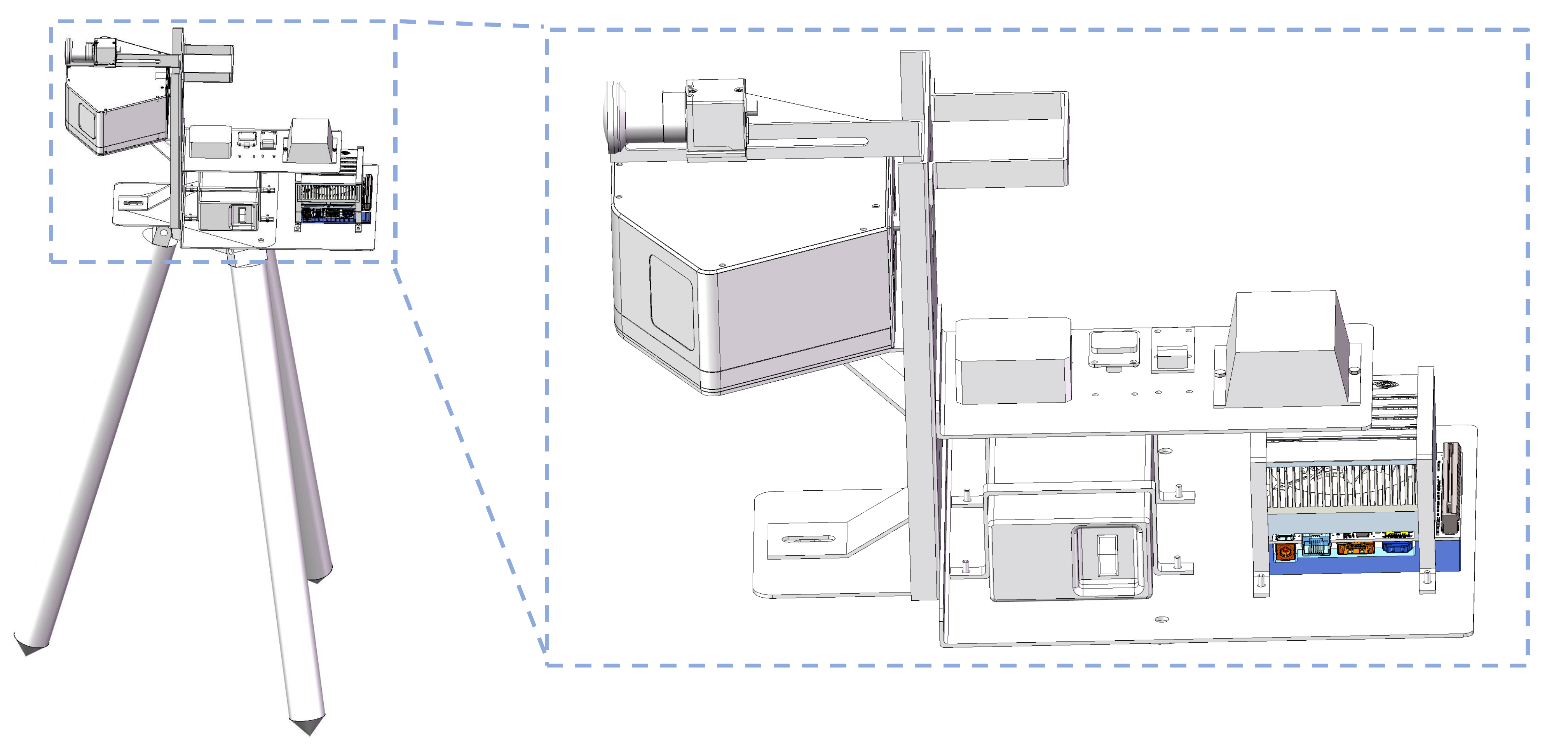}
    \caption{}
    \label{fig:assembly}
    \end{subfigure}
    \begin{subfigure}{.2\linewidth}
    \includegraphics[width=1\linewidth]{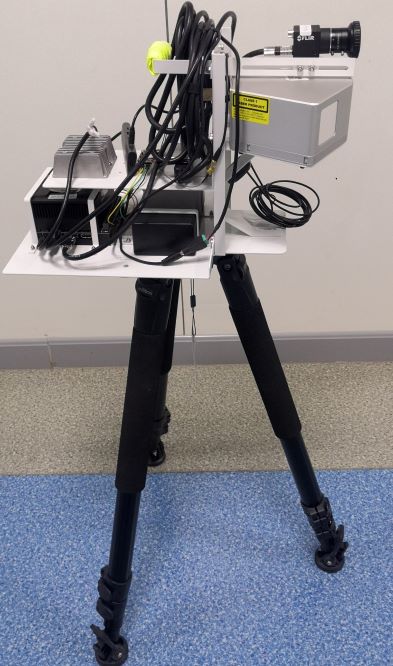}
    \caption{}
    \label{fig:setup}
    \end{subfigure}
    \caption{The designed model of slave node. Two parts can be disassembled and easily screwed together. A camera, a LiDAR, a Nvidia AGX Xavier and a battery are fixedly attached to the assembly body.}
    
\vspace{-.3cm}
\end{figure}

To improve the flexibility and scalability of our system, each slave node is made up of a support and a sensor assembly. As shown in Fig.~\ref{fig:setup}, the two parts can be disassembled and carried separately and easily screwed together at experiment sites. The support is modified based on a high-load-bearing tripod. As a result, maintaining, moving, and configuring each slave node is a breeze. We built a system of four slave nodes for evaluation while the 3D imaging system can freely expand or squeeze the node capacity due to the distributed multi-node network.

All sensors and other devices are fixedly placed in the assembly body, which is specifically designed to ensure the robustness of our system. In the front of the assembly, an RGB camera and a Livox Mid-100 LiDAR are mounted in the same horizontal position, with the camera located exactly above the LiDAR. As a result, the fields of view (FOV) of the LiDAR and the camera are comparable. Each Mid100 LiDAR can continuously scan the scene at 300,000 points per second for a horizontal view of $100\degree$ and a vertical view of $40\degree$ toward the front, providing accurate depth information. The camera and LiDAR are connected to an Nvidia AGX Xavier machine equipped on the sensor assembly. The local node machine participates as a storage unit and serves as a computational node. The Nvidia AGX Xavier machine provides computational efficiency comparable to RTX 1080, allowing lightweight models to be run on the slave node. For both the Nvidia AGX Xavier and the LiDAR, the slave nodes are powered by a mobile power supply with a 12V and 24V output. After a full charge, each slave node may operate continuously for more than three hours. In addition, the GPS module is employed as the accurate time source for time synchronization.

\section{Outdoor Scenes 3D Dataset}
\subsection{Overview}
We captured multi-view multi-modal data in different outdoor scenes using our system of 4 slave nodes. Each slave node is equipped with a FLIR industrial camera and a Livox mid100 LiDAR, and acquires a synchronized RGB image data stream and a point cloud data stream at 10 Hz. The time synchronization and spatial calibration are applied through the automatic spatio-temporal calibration paradigm described in Sec.~\ref{sec:calib}.
\begin{figure}[t]
    \centering
    \begin{minipage}{\linewidth}
        
\begin{tabular}{c|c}
\begin{subfigure}{0.49\linewidth}
    \begin{center}
    \includegraphics[width=\linewidth]{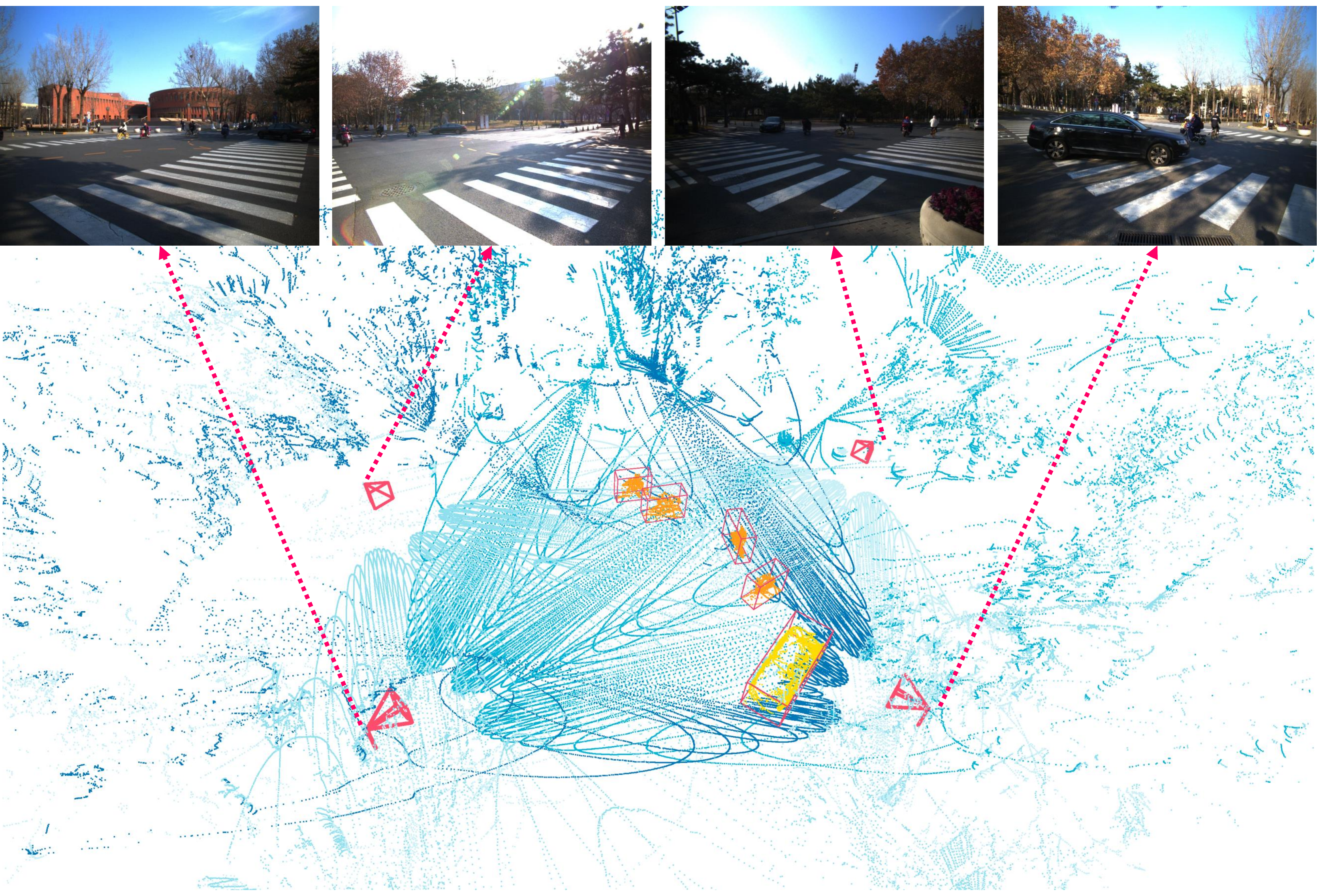}
    \caption{Crossroad scene.}
    \end{center}
    \end{subfigure}&
    \begin{subfigure}{0.49\linewidth}
    \begin{center}
    \includegraphics[width=\linewidth]{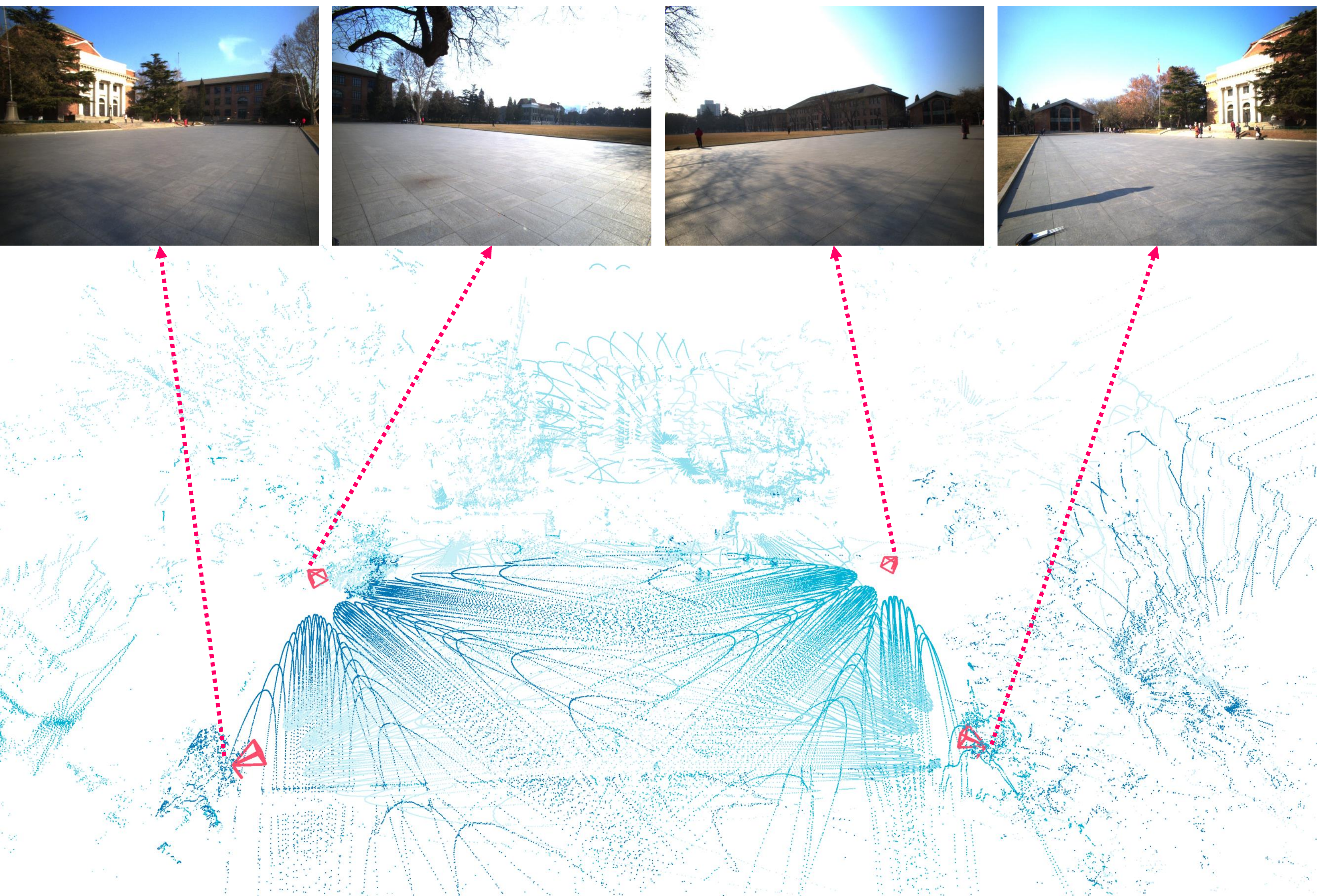}
    \caption{Large plaza scene.}\label{fig:plaza}
    \end{center}
    \end{subfigure}\\\hline
    
   \begin{subfigure}{0.49\linewidth}
    \begin{center}
    \includegraphics[width=\linewidth]{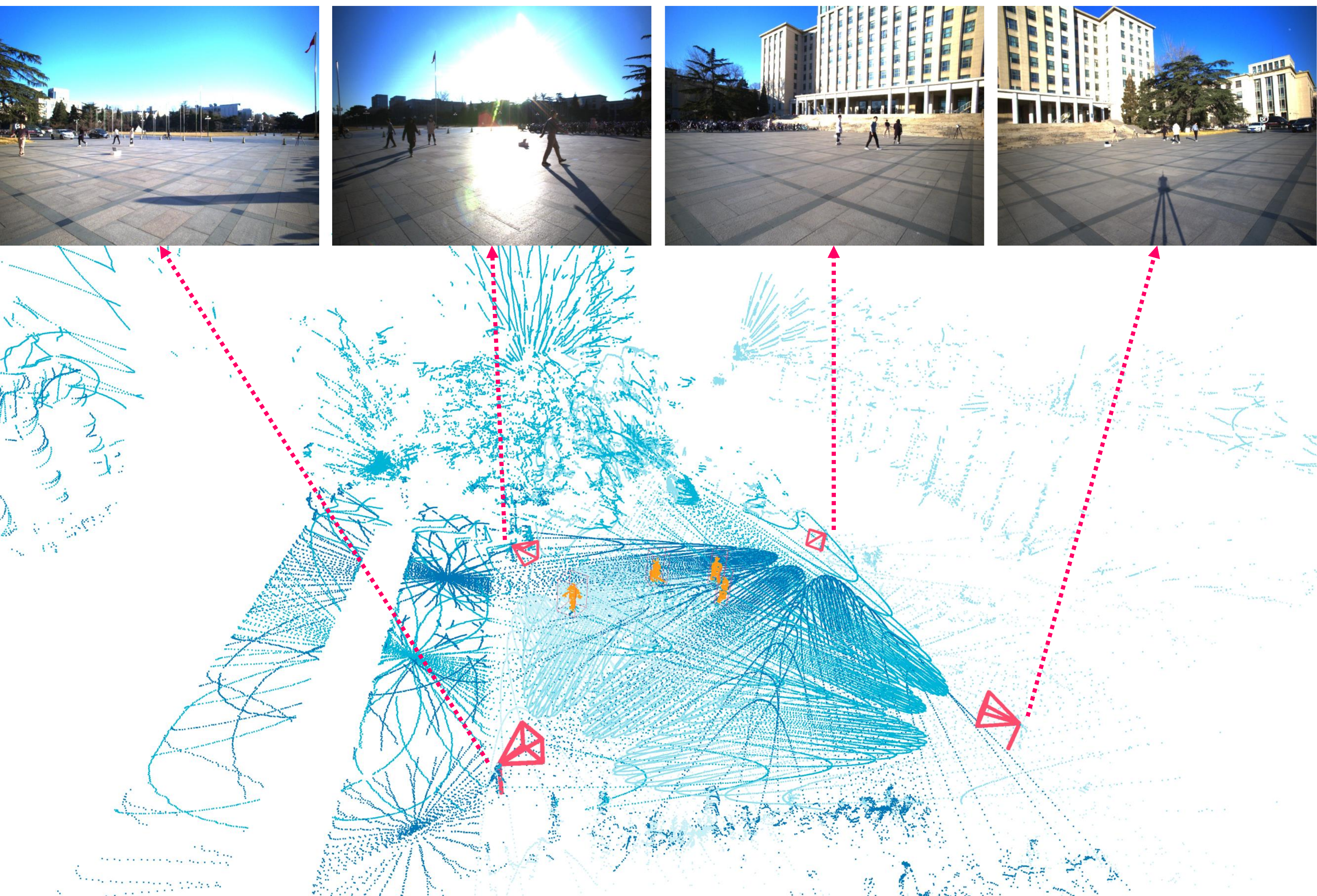}
    \caption{Multi-person scene at day.}
    \end{center}
    \end{subfigure}&
    \begin{subfigure}{0.49\linewidth}
    \begin{center}
    \includegraphics[width=\linewidth]{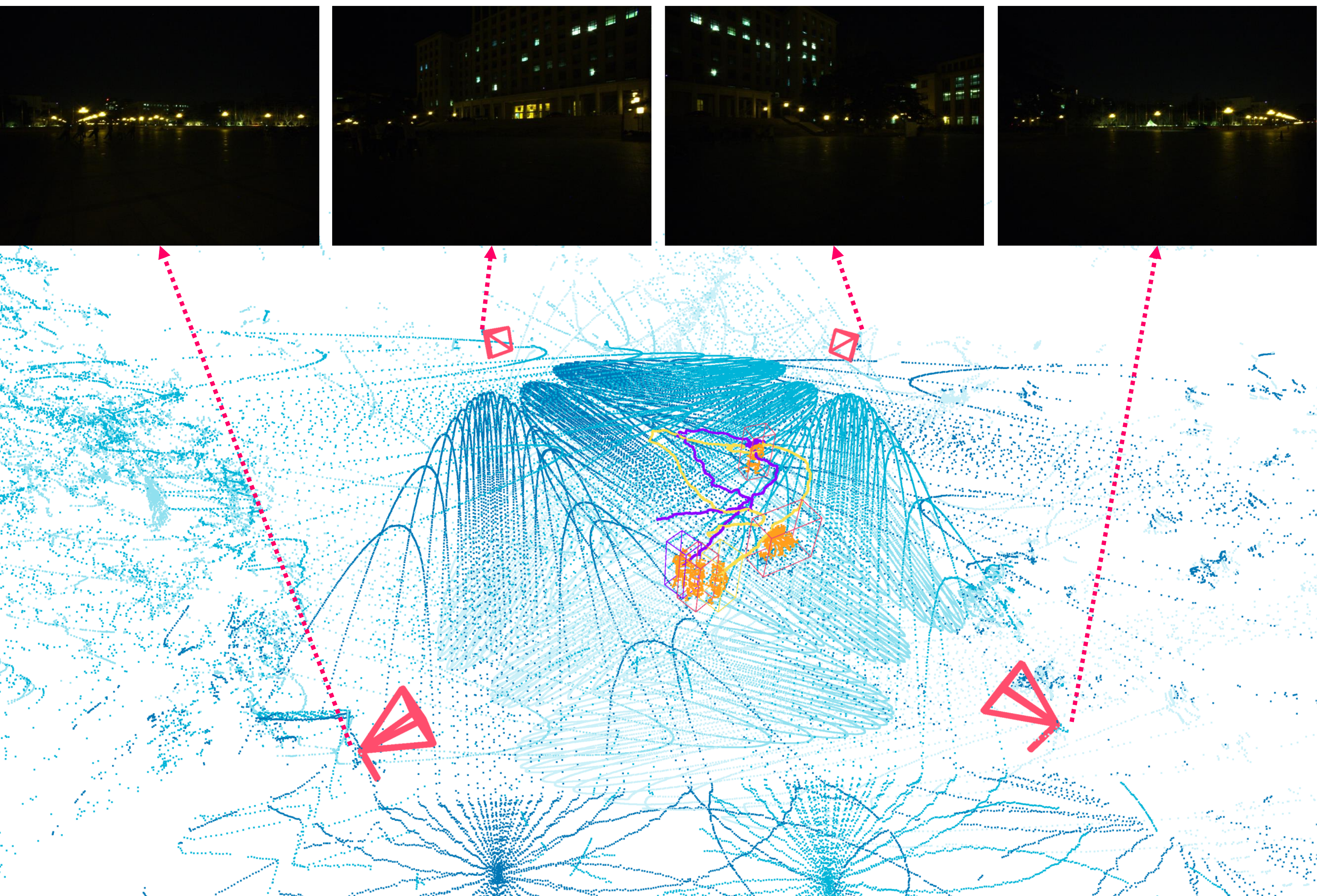}
    \caption{Multi-person scene at night.}\label{fig:night}
    \end{center}
    \end{subfigure}
\end{tabular}
    \end{minipage}
    \caption{Four acquisition scenes of our dataset. (a) is the crossroad scene with many vehicles and pedestrians. (b) is a square with a size of more than 50 meters. (c) and (d) is the multi-person scene in day and at night. In these scenes, we acquired over $10,000$ frames with high-precision synchronization and calibration from four slave nodes of our system.}\label{fig:scenes}
    
\vspace{-.3cm}
\end{figure}

The data acquisition took place in different scenes to evaluate the robustness of our system. We annotated the 3D bounding boxes and tracking ids of the targets in the scene using the annotation tool 3D BAT~\cite{DBLP:conf/ivs/3dbat}. A total of more than $10,000$ frames were obtained, with $8,400$ of them being labeled. We briefly describe the overview of each scene below, which is demonstrated in Fig.~\ref{fig:scenes}. All collected data and annotations have been made publicly available.

\subsection{Acquisition Scenes}
    

\subsubsection{Crossroad Scene}
\begin{figure}
    \centering
\begin{minipage}{0.9\linewidth}
    
    \centering
    \begin{subfigure}{.32\linewidth}
    \includegraphics[width=\linewidth]{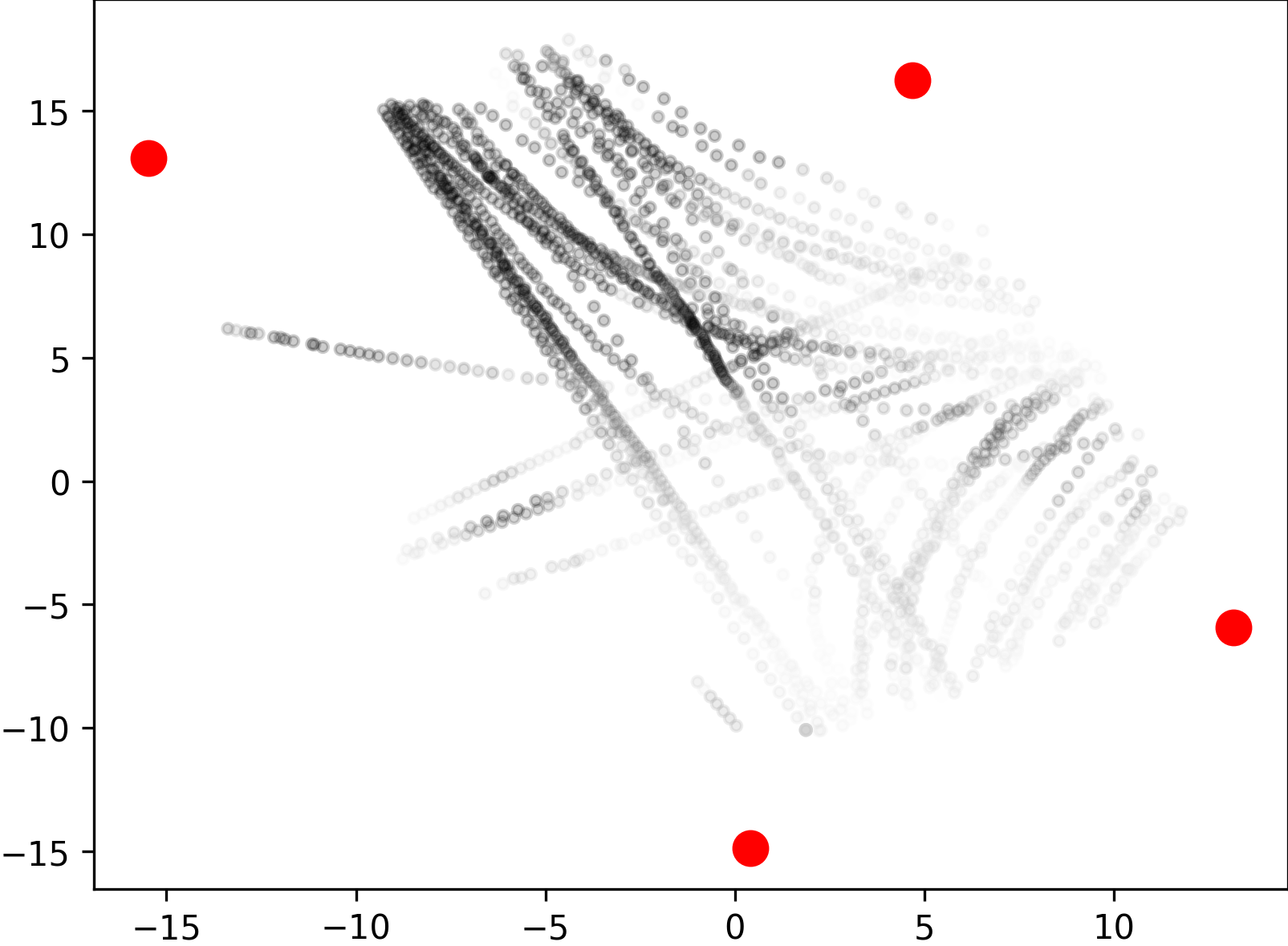}\\
    \includegraphics[width=\linewidth]{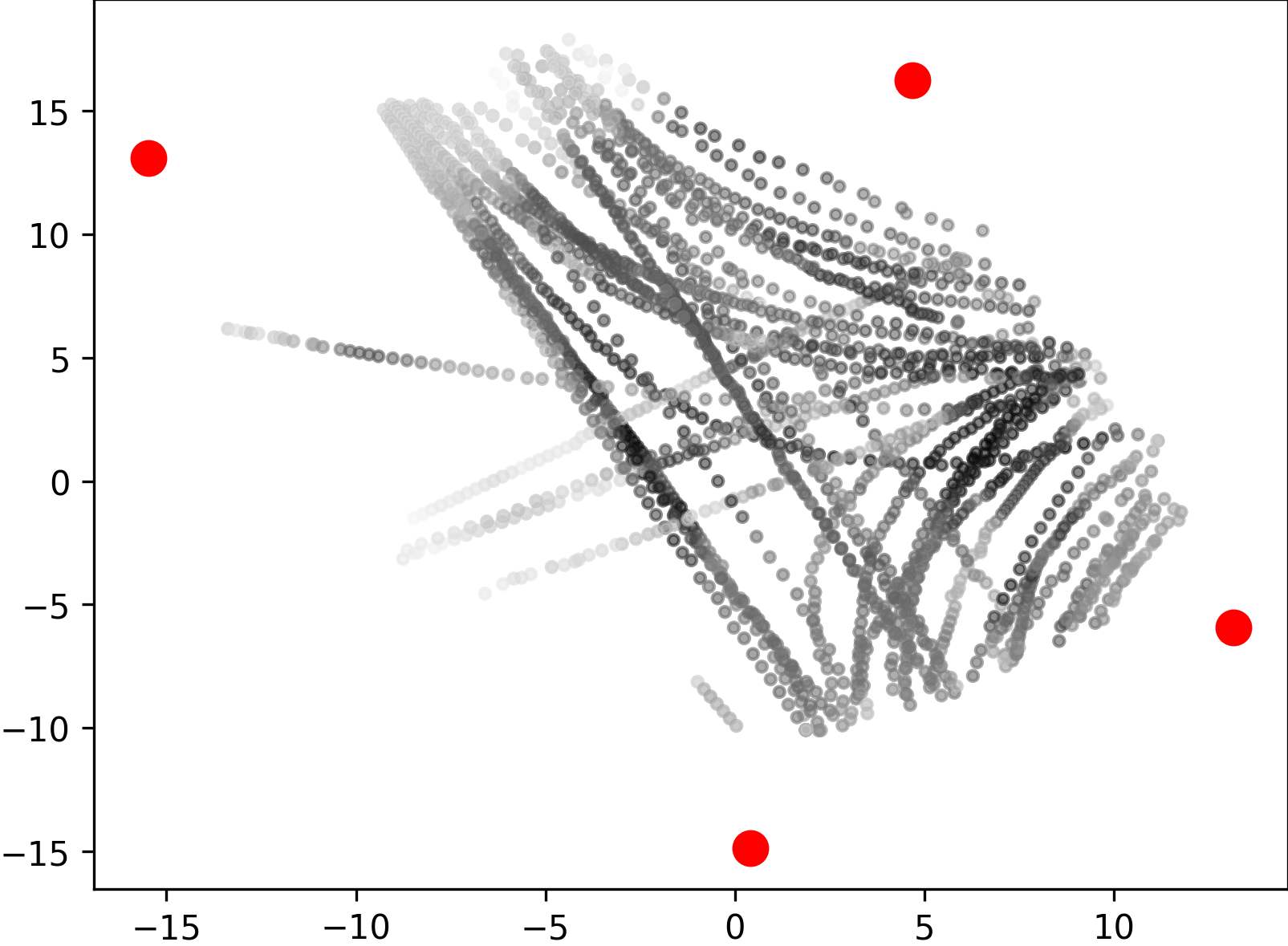}
    \caption{Car}
    \end{subfigure}
    \begin{subfigure}{.32\linewidth}
    \includegraphics[width=\linewidth]{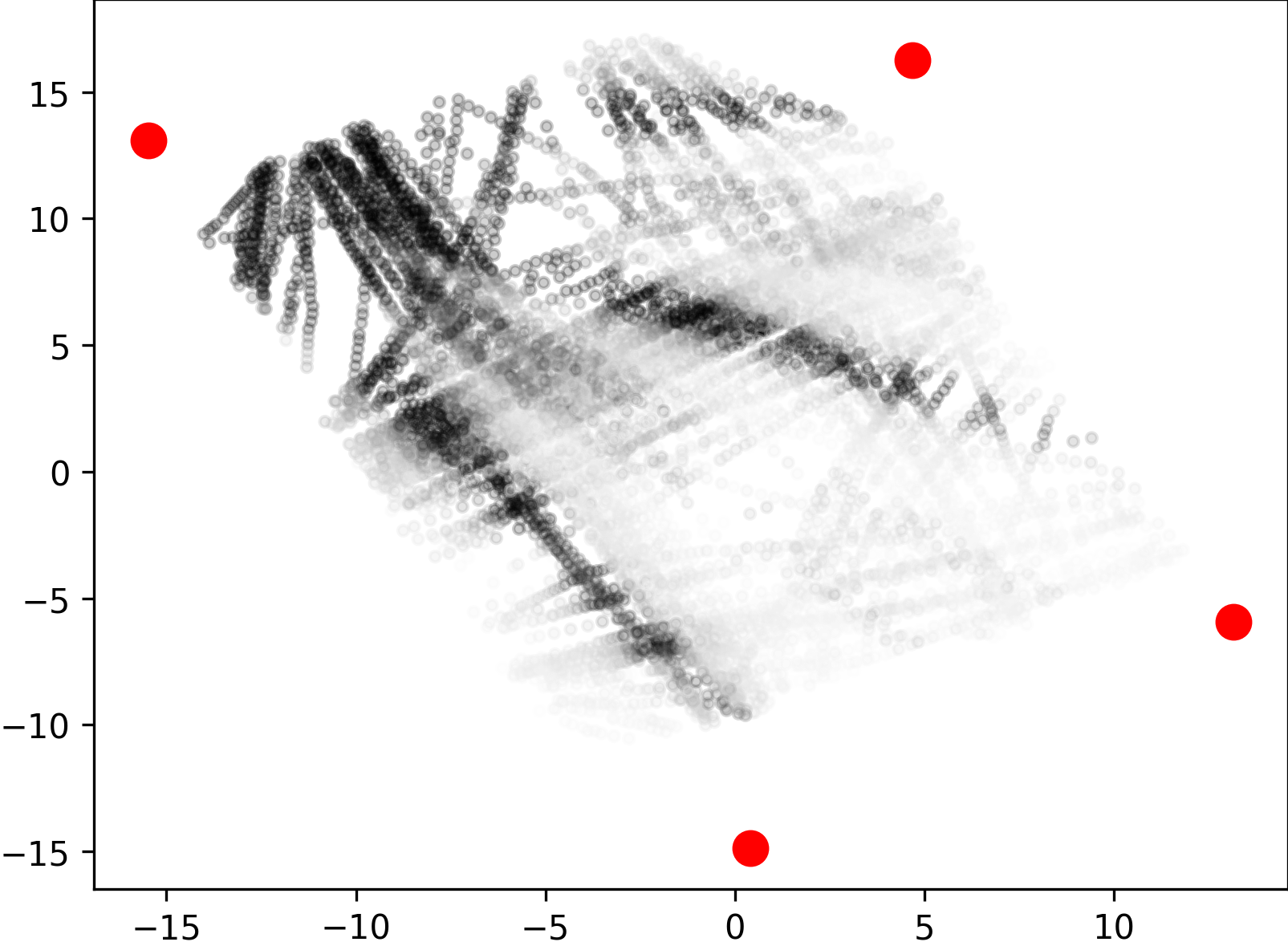}\\
    \includegraphics[width=\linewidth]{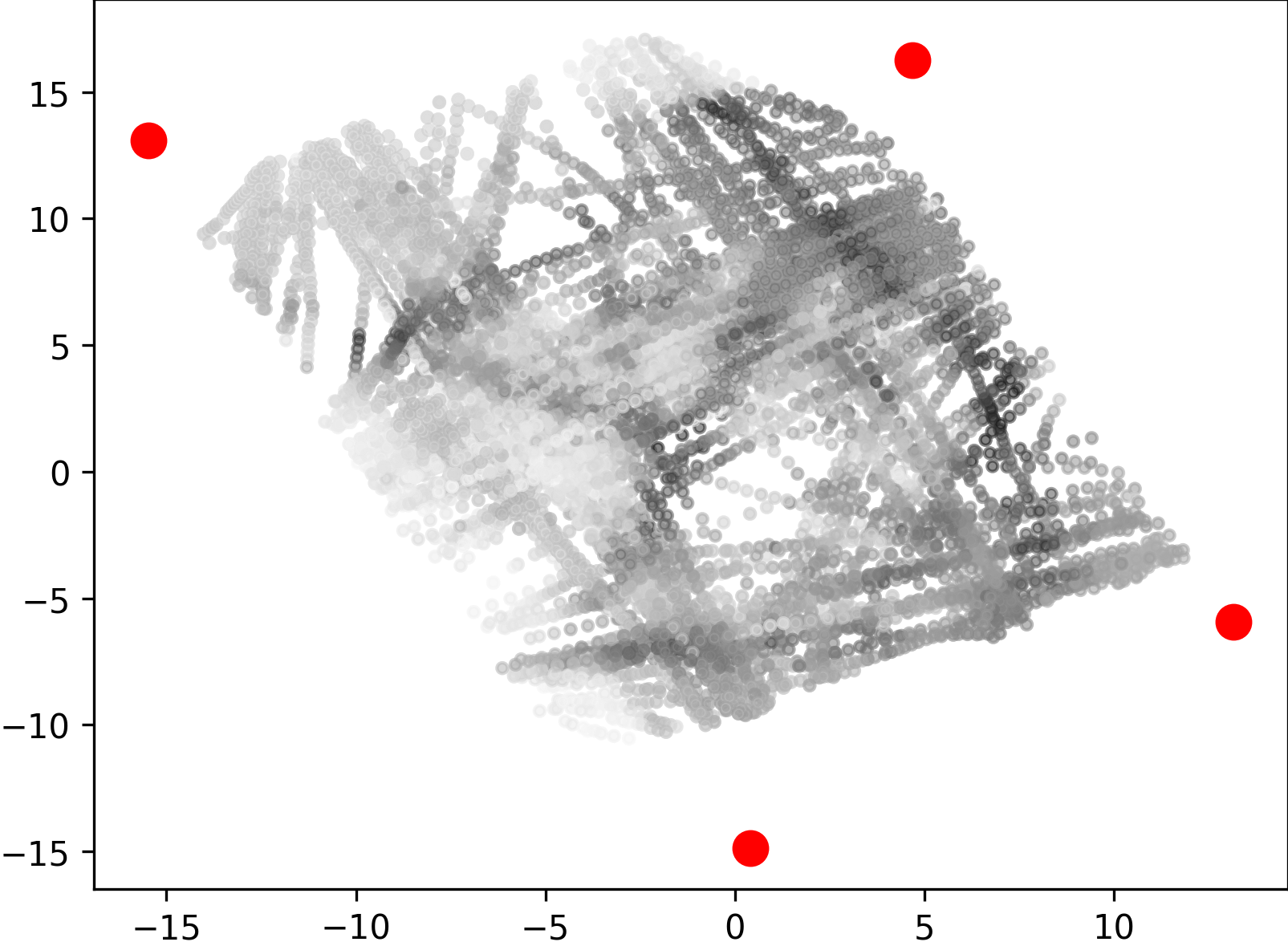}
    \caption{Cyclist}
    \end{subfigure}
    \begin{subfigure}{.32\linewidth}
    \includegraphics[width=\linewidth]{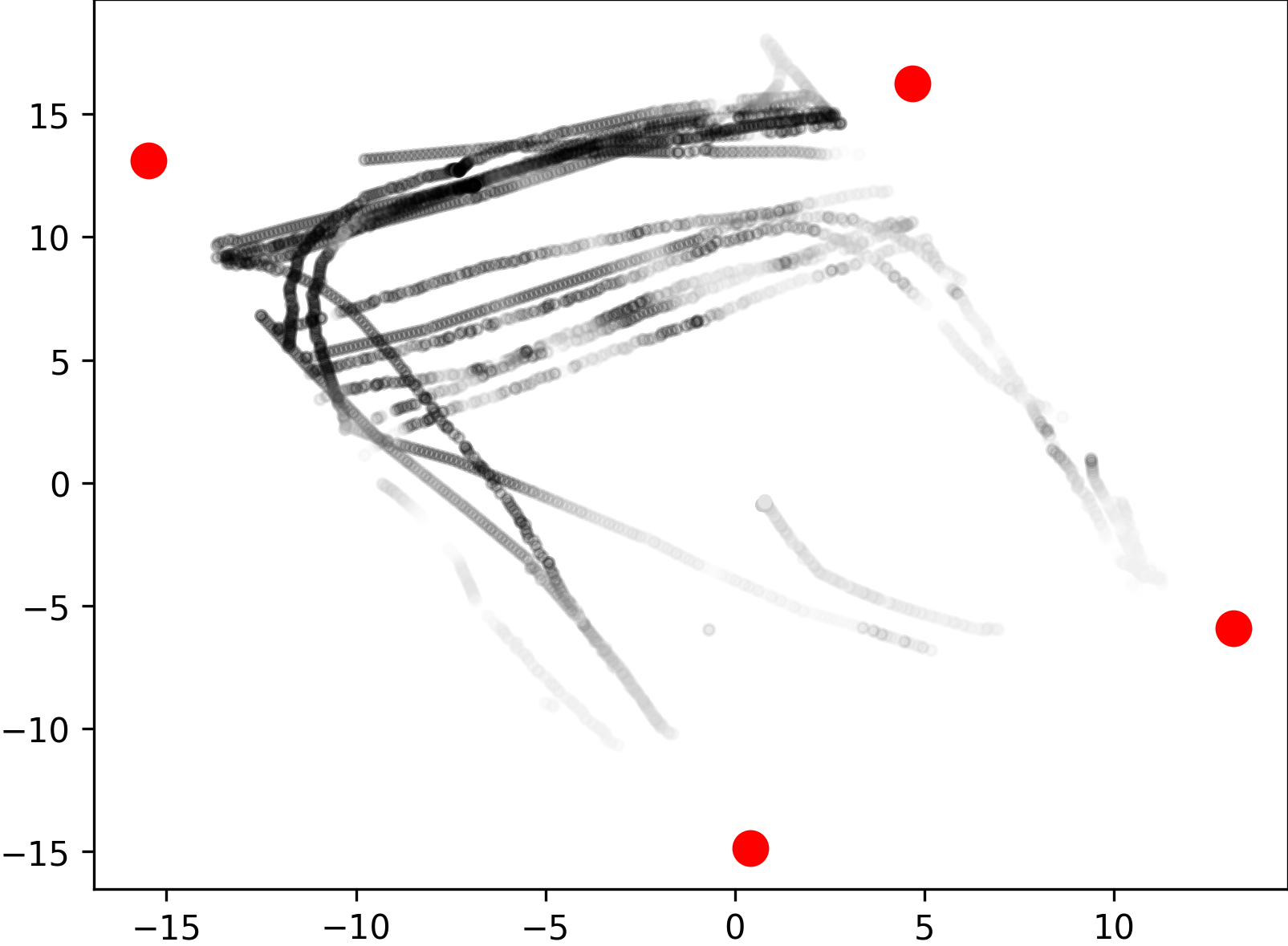}\\
    \includegraphics[width=\linewidth]{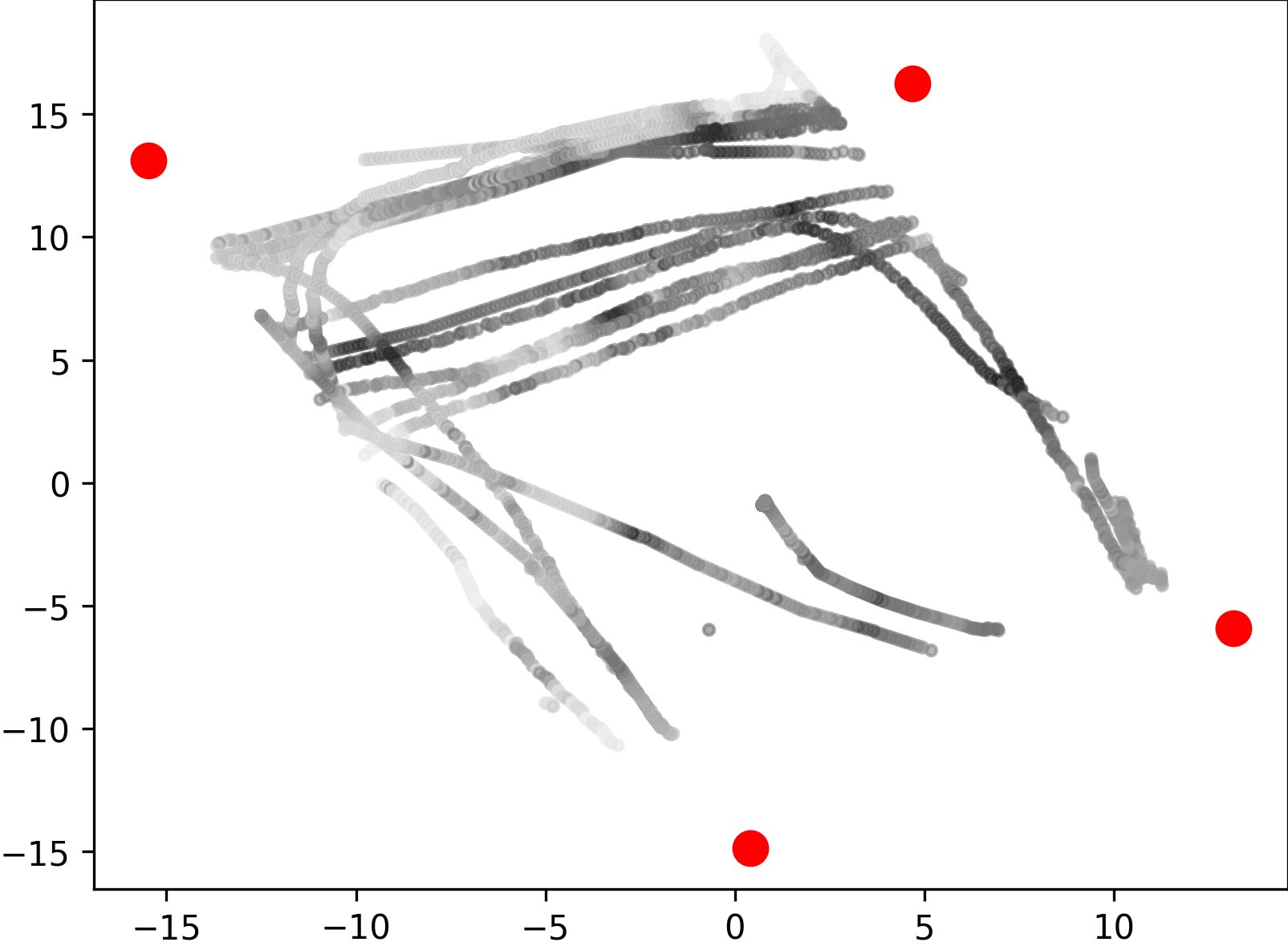}
    \caption{Pedestrian}
    \end{subfigure}
\end{minipage}
    \caption{The distribution of objects in the crossroad scene for single view (top) and four views (bottom). All objects are depicted as scatters in bird's-eye view, with the darker the color, the more 3D points in the object. A red point represents a slave node. The objects close to slave node have more points for the single view.}\label{fig:occupy}
\end{figure}

We collected 3D dynamic data of traffic scenes by 4 slave nodes set up at the four corners of a street crossroad, where many vehicles and pedestrians moved across the scene. We provided $4,000$ frames with 3D bounding box annotations of cars, cyclists, and pedestrians in the central area of the crossroad, which were used for the evaluation of 3D object detection methods. Unlike other street scene datasets, our dataset contains a large number of cyclists and pedestrians on the campus. Cars, cyclists, and pedestrians are distributed in a $1:4:1$ ratio, which amounts to a total of more than $15,000$ bounding boxes with track ids. All cars, cyclists, and pedestrians are depicted in Fig.~\ref{fig:occupy} as scatters in bird's-eye view, with the darker the color, the more points in the object. We can see that using multi-view LiDARs enables uniform coverage of 3D space when compared with single-view LiDAR, which is important for 3D object detection and tracking in large scenes.

\subsubsection{Multi-person Scene}
Some applications, such as football game analysis, require methods with long-term tracking capabilities, as opposed to short-term tracking of targets in open scenes.
To evaluate the performance of long-term objects tracking, we collected data of multiple persons moving for a long time in the plaza and rarely leaving, for 1200 frames in the day and 1200 frames at night. The persons are free to move and interact with others, which results in a lot of occlusions and intersected trajectories, causing great difficulties for single-view data. Furthermore, we repeated the collection at night, under very poor lighting conditions. As shown in Fig.~\ref{fig:night}, while RGB cameras can hardly produce high quality data even with increased exposure time, the multi-view point clouds from LiDAR sensors are completely unaffected, which indicates the robustness of our multi-modal system. All of the frames were annotated with 3D boxes and track ids for 3D multi-object tracking. A subset of the ground-truth trajectories are depicted in Fig.~\ref{fig:track} with colored lines.

In a plaza with a size of over 50 meters, we also utilized our system to collect multi-view multi-modal data, which verified the applicability of our system in large outdoor scenes. This dataset contains 4800 frames and has yet to be annotated, which is shown in Fig.~\ref{fig:plaza}. Our system's synchronization and calibration performed reliably in the difficult large-scale scenes without wired connections.

\subsection{System Reliability}
\begin{figure*}[h]
    \centering
\begin{minipage}{0.44\linewidth}
    \centering
    \begin{subfigure}{.49\linewidth}
    \includegraphics[width=\linewidth]{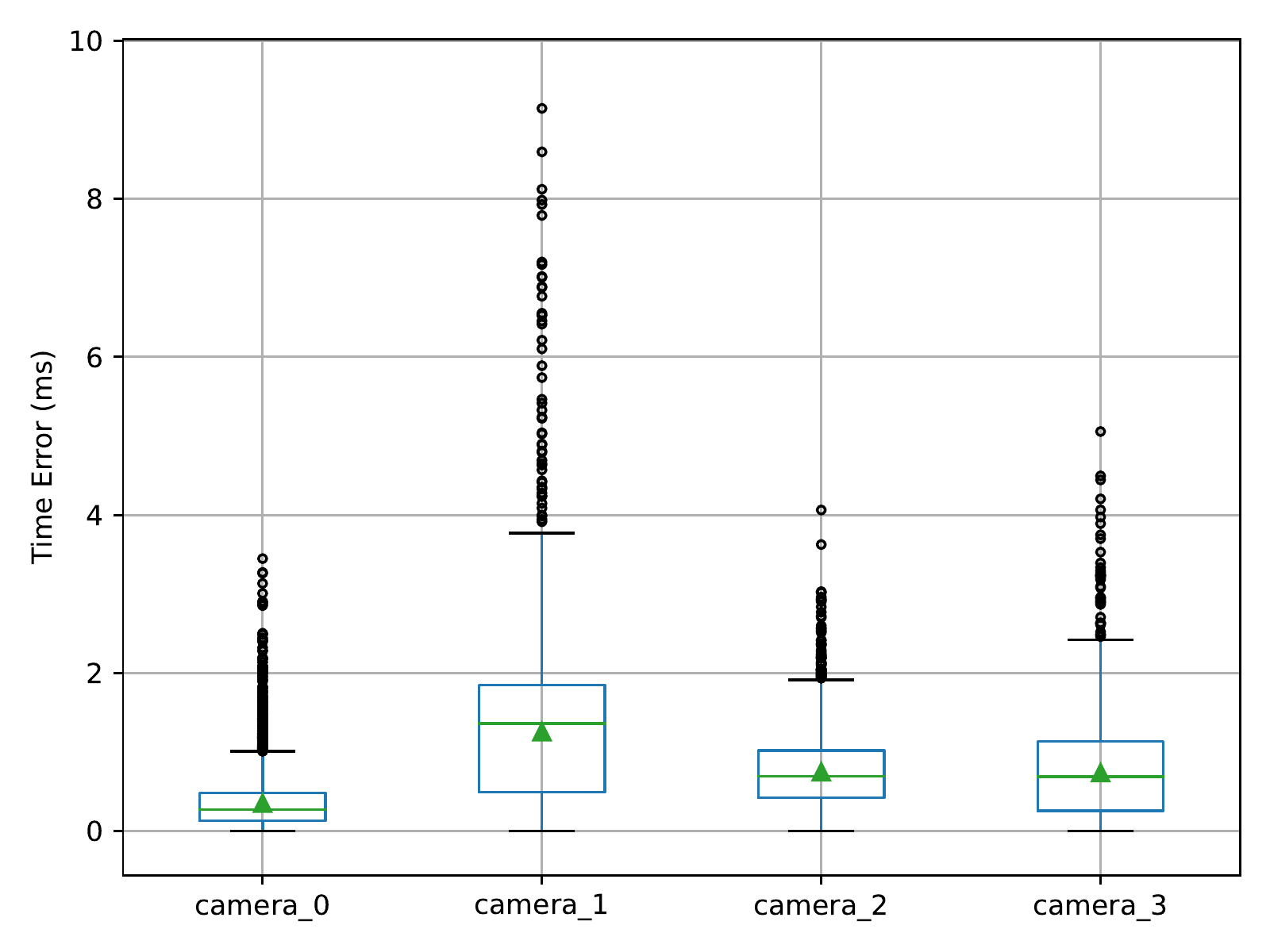}
    \caption{Time error of images.}
    \end{subfigure}
    \begin{subfigure}{.49\linewidth}
    \includegraphics[width=\linewidth]{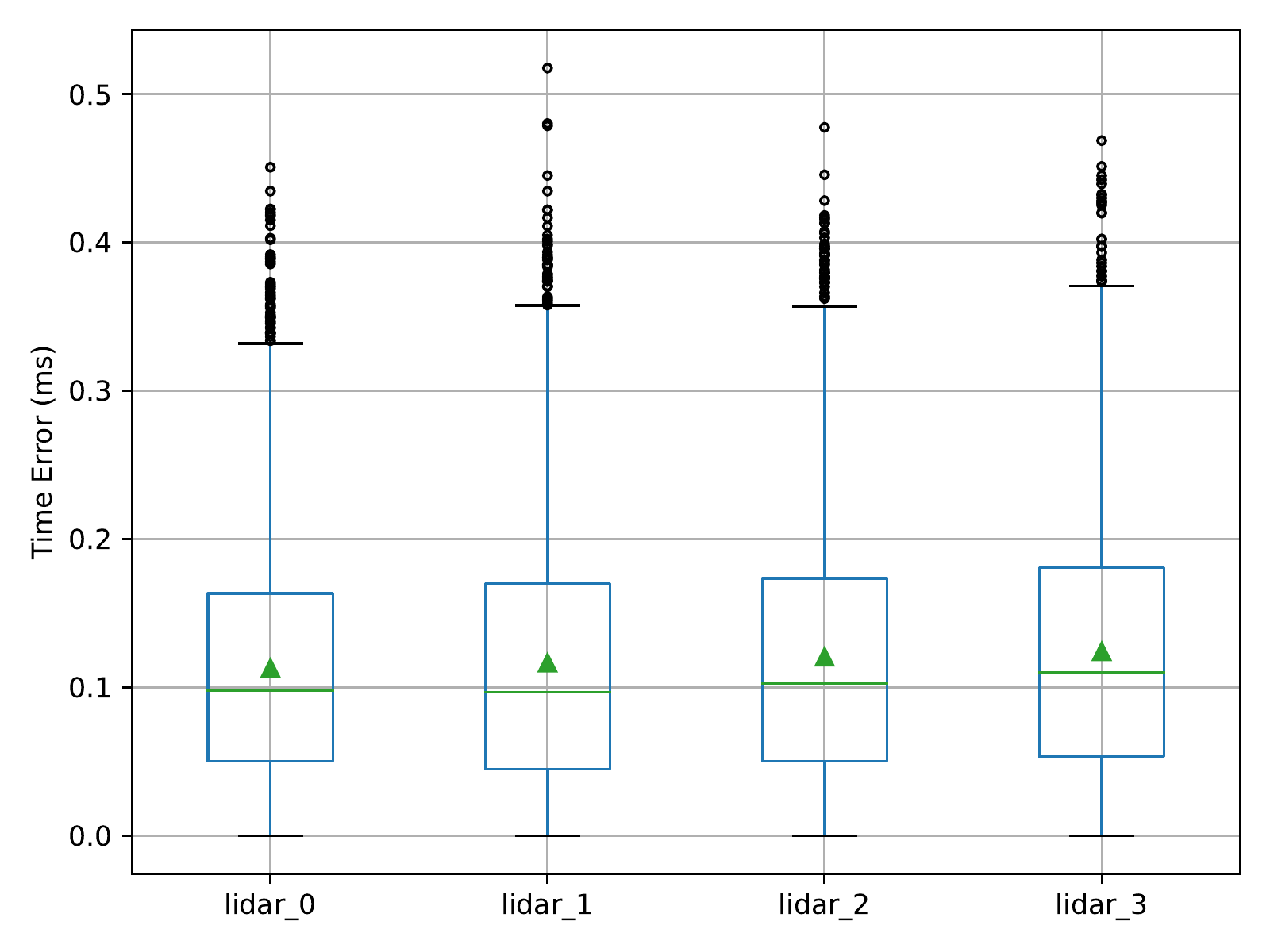}
    \caption{Time error of point clouds.}
    \end{subfigure}
    \caption{The time error of image and point cloud data for four nodes in the acquired dataset.}\label{fig:time_error}
\end{minipage}
    \begin{minipage}{0.55\linewidth}
\centering
        \captionof{table}{The object detection performance of baselines using the multi-view point clouds.}\label{tab:baseline}
\begin{tabular}{l|c|c|c|c}
\hline\hline
             & Pedestrian & Cyclist & Car & Overall \\ \hline
SECOND~\cite{DBLP:journals/sensors/second} & 57.91 &   89.96  &  68.89  & 72.26          \\ 
PointPillars~\cite{DBLP:conf/cvpr/pointpillars} &50.53  &  89.55 &  74.09 &  71.39 \\ 
3DSSD~\cite{DBLP:conf/cvpr/3dssd}  & 47.59    &  80.59  & 77.05&  68.41    \\ \hline
\end{tabular}
    \end{minipage}
    
\vspace{-.1cm}
\end{figure*}

During the capture of our dataset in different scenes, the accuracy of synchronization and calibration was assessed to evaluate our system's reliability. As all four node’s system clocks were synchronized by GPS modules to have an error less than 1 us~\cite{timesync}, which is negligible in comparison to total time error, we estimated the synchronization error as the difference in the data's timestamp. The frame's reference time was determined by the average timestamp of image data from four nodes. As shown in Fig.~\ref{fig:time_error}, we can see that four point cloud data streams were synchronized perfectly with less than 1 ms error 
in favor of multi-view data fusion. The time error in image data is more uncontrollable due to the software trigger mode, though the mean error of 1 ms is acceptable for the large-scale outdoor scenes. 
To evaluate the calibration of different nodes in the system, we annotated 20 corner points evenly distributed throughout each scene in the point cloud. Then we calculated the mean of point-point projection error after
applying the transformation as the evaluation metric. The mean point-point projection error is $0.029$ m for multi-person scene and $0.023$ m for the crossroad scene. In addition, we manually calibrated the extrinsic parameters between Livox LiDARs and cameras following the Livox official instructions~\cite{extrinsic}. The mean reprojection error for four nodes is $2.98$ pixels, which can be improved in the future with more annotated corner points and stronger camera-lidar calibration methods.

Through the dataset acquisition at multiple different sites, we also established the reliability of the system. Our system worked effectively both during the day and at night at crossroads with complex surrounds and open squares. Our system prefers a robust GPS signal to synchronize slave nodes quickly. For a successful calibration, the slave nodes should be placed within 200 meters of some static calibration objects like buildings, traffic signs, or trees, which are quite common. In fact, the proposed system works well in all the places we tested on our campus (not just the places of the three datasets).

\section{Experimental Results}
The experimental results on our dataset are presented in this section. Our collected data, unlike other multi-view datasets, includes the multi-view point cloud from multiple LiDARs, which complements the target's point cloud from all directions. In this regard, we conducted experiments of 3D object detection and 3D multi-target tracking on the multi-view point cloud sequence data. More experimental results are available in the supplementary materials.
\subsection{Evaluation Metrics}
For 3D object detection, we assessed the performance using the established average precision (AP) metric as described
in KITTI~\cite{DBLP:conf/cvpr/kitti}. We regarded detections as true positives to overlap by more than $70\%$ for cars and $50\%$ for cyclists and pedestrians.
Regarding multiple object tracking, We used MOTA, MOTP, IDS, FRAG and FN described in
CLEAR~\cite{DBLP:journals/ejivp/clear} as main tracking metrics.

\subsection{3D Object Detection}
\subsubsection{Baselines}
We ran a representative set of point cloud based 3D detection methods on this dataset, and the results are shown in Tab.~\ref{tab:baseline}. We used the first 2000 frames of the crossroad scene dataset as the training set and the last 2000 frames as the validation set. We reproduced and evaluated the 3D detection methods of PointPillars~\cite{DBLP:conf/cvpr/pointpillars}, SECOND~\cite{DBLP:journals/sensors/second} and 3DSSD~\cite{DBLP:conf/cvpr/3dssd} on our multi-view point cloud dataset. As shown in Tab.~\ref{tab:baseline}, most of the methods achieved good results on our dataset. The precision of pedestrians and cars is comparable to results in KITTI benchmark~\cite{KITTI}, while our detection performance of cyclists is higher due to massive of cyclists in the scene. It should be mentioned that KITTI's annotations are based on a single LiDAR, and heavily occluded objects are not labeled. On the contrary, there are almost no objects unlabeled due to occlusion in our data.

\subsubsection{More views improve performance}
The multi-view imaging system proposed in this paper employs multiple LiDARs that scan from all directions to obtain the point cloud of targets. Compared to employing simply a single view, integrated point cloud considerably reduces the impact of occlusions and improves the accuracy of the target's location and size. In this regard, we conducted experiments on the effect of the number of views for 3D object detection.
PointPillars~\cite{DBLP:conf/cvpr/pointpillars} with the same configure is considered as the baseline method to test the point clouds for different groups of views respectively. For the fairness of the experiment, we performed a point cloud integration of multiple original frames for the single-view group and double-view group to ensure that they have the same amount of points per frame as all four views. Integration is implemented by fusing all point clouds and appending a time index to each point as temporal information.

\begin{table*}[t]
\centering
\caption{The object detection performance of using different group of views. The detection results for pedestrians, cyclists and cars are evaluated by $AP_{50}$, $AP_{50}$ and $AP_{70}$. Overall results are calculated by the mean AP of three classes.} \label{tab:view}
\begin{tabularx}{.87\linewidth}{l|C{1cm}C{1cm}C{1cm}C{1cm}|C{1.45cm}C{1.45cm}C{1.45cm}|c}
\hline
\hline
Group                         & view0 & view1 & view2 & view3 & Pedestrian & Cyclist & Car & Overall \\ \hline
\multirow{2}{*}{Single view}  &  \checkmark & & & & 28.80 &  85.43   & 46.23   & 53.49 \\ 
                              &        &        & \checkmark & &  32.51   &  64.90  & 40.89 & 46.10      \\ \hline
\multirow{2}{*}{Double views} &  \checkmark &  & \checkmark  &  &  37.37   &  88.38 & 57.29& 61.02   \\  
                              &        &  \checkmark &  &  \checkmark  &  45.55  & 85.38  & 61.60 & 64.18 \\ \hline
\multirow{2}{*}{Triple views} &   \checkmark &  \checkmark & \checkmark  &  &41.18 &\bf 89.76 &73.04 & 67.99  \\ 
                              &        & \checkmark &  \checkmark & \checkmark  &37.65  &87.24  &  70.06  & 64.98   \\ \hline
Four views   &      \checkmark &  \checkmark & \checkmark  &  \checkmark  & \bf 50.53  &  89.55 & \bf 74.09 & \bf 71.39   \\\hline 
\end{tabularx}
\end{table*}

\begin{figure}
    \centering
    \begin{subfigure}{.32\linewidth}
    \includegraphics[width=\linewidth]{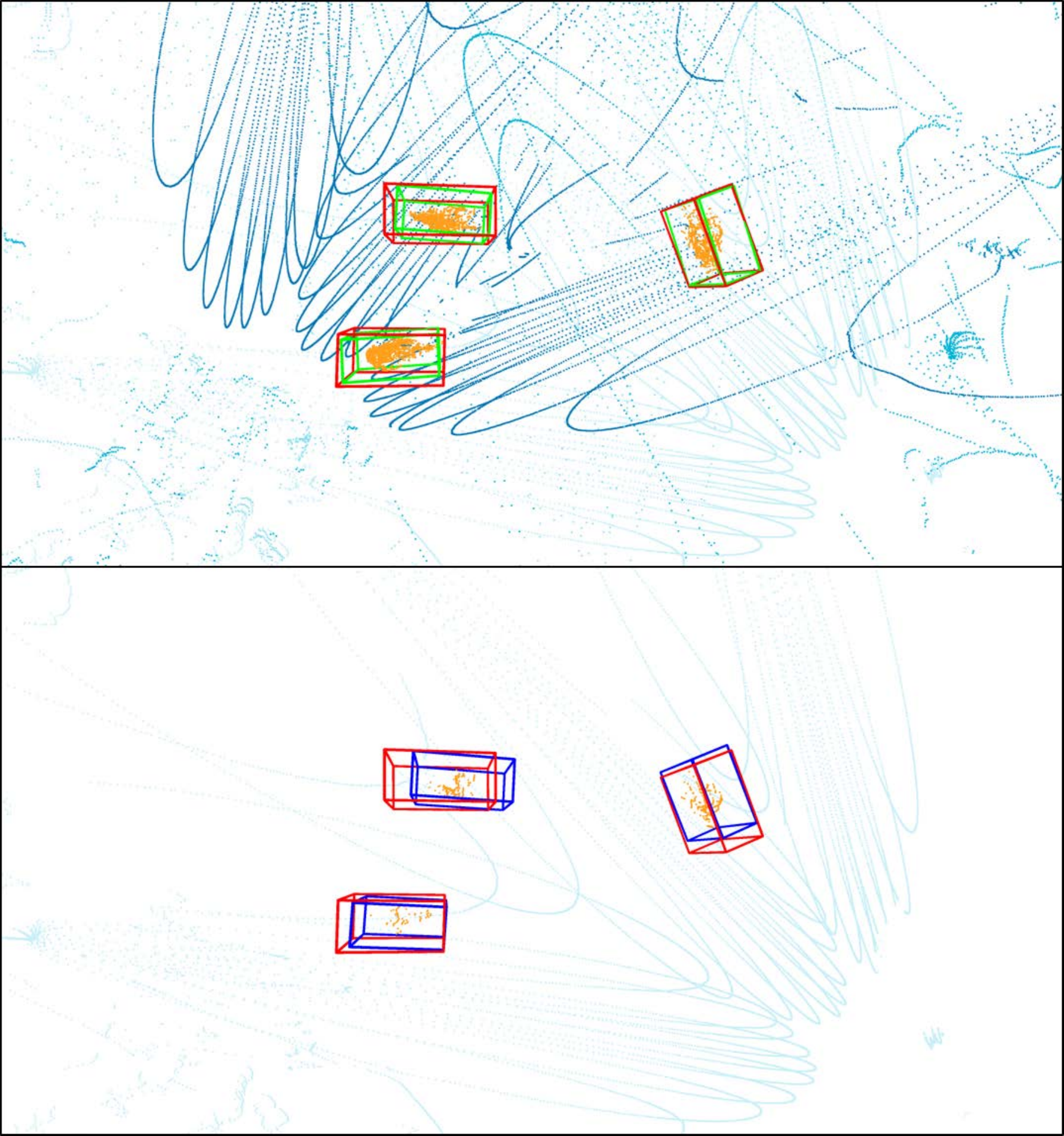}
    \caption{}\label{fig:bias}
    \end{subfigure}
    \begin{subfigure}{.32\linewidth}
    \includegraphics[width=\linewidth]{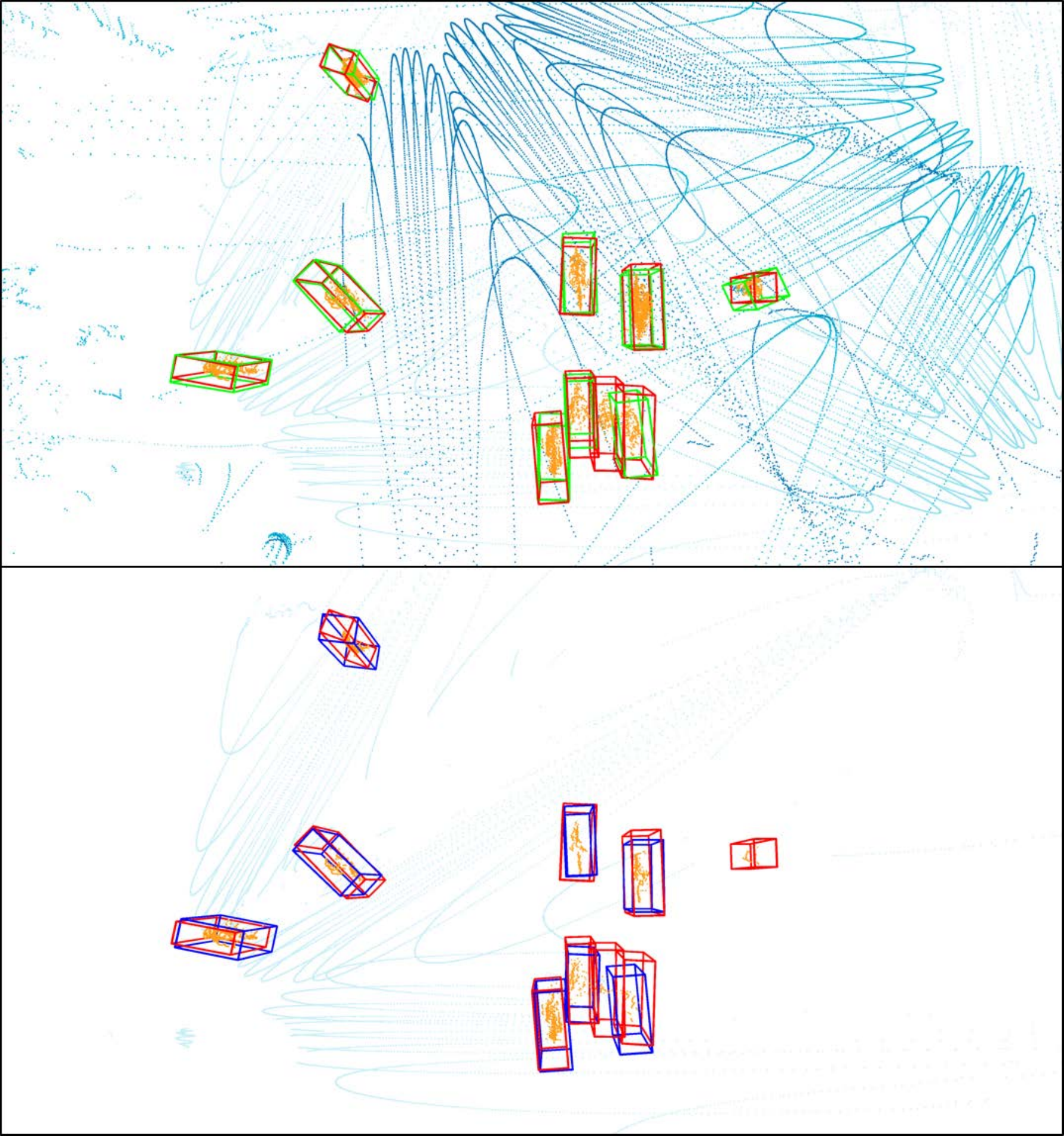}
    \caption{}\label{fig:miss1}
    \end{subfigure}
    \begin{subfigure}{.32\linewidth}
    \includegraphics[width=\linewidth]{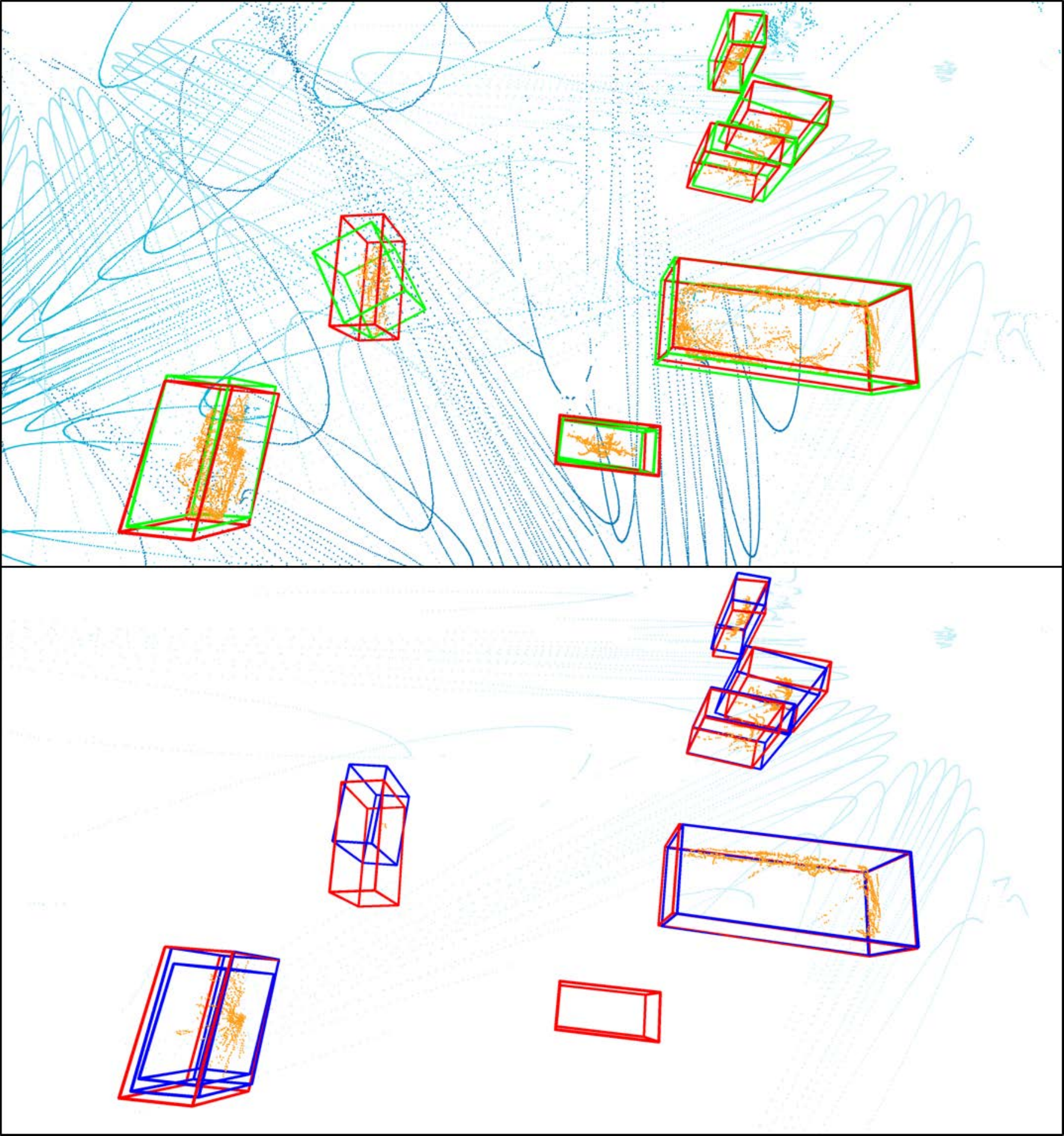}
    \caption{}\label{fig:miss2}
    \end{subfigure}
    \caption{Qualitative results of a single view and all four views. The results of the multi-view model are on top, with green boxes, and the results of the single-view model are on the bottom, with blue boxes. Red boxes represent the ground-truth bounding boxes.}
    \label{fig:my_label}
    
\vspace{-.3cm}
\end{figure}
The experimental results in Tab.~\ref{tab:view} show that the more views used, the higher the detection accuracy. The best detection performance is achieved when all four views are used, especially for the difficult pedestrian class. Using view 0, 1 and 2 got a equivalent performance of cyclists and cars as all four views, since the occlusion of these two classes is not severe. The qualitative results are shown in Fig.~\ref{fig:my_label}, with the multi-view model's results on the top and the single-view model's results on the bottom. The ground-truth bounding boxes are displayed by red bounding boxes. As we can see, the point cloud of multiple views effectively eliminates the impair of occlusion between objects, improving the recall of detection. On the other hand, occlusion has a significant impact on single-view detection results. The pedestrian in the right of Fig.~\ref{fig:miss1} is occluded 
with only a few points left and the cyclist in the bottom of Fig.~\ref{fig:miss2} is completely invisible for the single view. As a result, the occluded objects are missed in single-view detections but captured by the multi-view model. Furthermore, since the shape of the object obtained from multiple views is more complete, the exact position and size of the object can be more reliably inferred using the point cloud of multiple views. When using the point cloud of a single view, the predicted position will be more biased toward the dense location of the points, as shown in Fig.~\ref{fig:bias} and Fig.~\ref{fig:miss2}.

\subsubsection{Late fusion and early fusion}
There are two intuitive approaches to handle the point cloud of multiple views: one is to explicitly integrate all of them and feed them to the model for prediction, which is known as early fusion; the other is to predict locally for each view and subsequently fuse the detection results, which is known as late fusion.
\begin{table*}[t]
\centering
\caption{The object detection performance of early fusion and late fusion approaches.} \label{tab:fusion}
\begin{tabular}{l|cc|cc|cc|c}
\hline\hline
\multirow{2}{*}{Method} & \multicolumn{2}{c|}{Pedestrian} & \multicolumn{2}{c|}{Cyclist} & \multicolumn{2}{c|}{Car} & \multirow{2}{*}{Overall} \\ \cline{2-7}
                        &\ \  $AP_{50}\ \ $      & \ \ $AP_{25}\ \ $      & \ \ $AP_{50}\ \ $     & \ \ $AP_{25}\ \ $    & \ \ $AP_{70}$\ \    & \ \ $AP_{50}$\ \   &                          \\ \hline
view 0                  & 28.80          & 79.59          & 85.43         & 90.03        & 46.23       & 86.78      & 53.49                    \\
view 1                  & 17.57          & 59.41          & 75.56         & 87.50        & 47.15       & 89.52      & 46.76                    \\
view 2                  & 32.51          & 61.18          & 64.90         & 85.62        & 40.89       & 80.13      & 46.10                    \\
view 3                  & 31.87          & 51.13          & 65.16         & 86.31        & 37.46       & 79.36      & 44.83                    \\ \hline
NMS Fusion              & 33.78          & 80.40          & 85.98         & 90.07        & 47.01       &  89.81      & 55.59                    \\
Average Fusion          & 42.82          & \bf86.23          & 88.74         & 90.39        & 59.85       & 90.04      & 63.81                    \\ \hline
Early Fusion          & \bf 50.53          & 81.65          & \bf89.55         & \bf98.26        & \bf74.09       & \bf90.61      & \bf71.39                    \\ \hline
\end{tabular}
\end{table*}
We evaluated two methods of late fusion: non-maximum suppression (NMS) fusion and average fusion~\cite{solovyev2021weighted}. Bounding boxes from four single views are obtained and combined together in both methods, and those that overlap by $0.1$ were deemed to belong to the same object. NMS fusion selected the box with the highest score as the object's final detection. For average fusion, the results for the same object were averaged to reach the final bounding box.

The final experimental results are shown in the Tab.~\ref{tab:fusion}, which reveal that the early fusion approach with direct integration has a greater detection accuracy than late fusion due to less information loss. By assembling information from four views, late fusion approaches outperform single-view approaches. Average fusion performs better by refining the location and size of boxes than NMS fusion.
Despite a slightly poorer detection performance, late fusion is load-balanced by using the local computation unit of each node and only transmits the detection results to the master node,  reducing the burden on data transmission in practical applications.
\subsection{3D Multi-object Tracking}
We conducted experiments on 3D multi-object tracking for long-term persistent targets using multi-person dataset, in which four pedestrians moved and interacted freely to create long-term trajectories. We utilized PointPillars~\cite{DBLP:conf/cvpr/pointpillars} to get the detection results, and then used AB3DMOT~\cite{DBLP:conf/iros/ab3dmot} to obtain the final tracking results. Due to the accurate depth of multi-view point clouds, our tracking results obtain a better performance than multi-view datasets~\cite{DBLP:conf/cvpr/Wildtrack,DBLP:conf/icmla/EPFL} in similar scenes.
\begin{figure*}
\begin{minipage}{.7\textwidth}
\centering
\captionof{table}{Tracking results on multi-person dataset.}\label{tab:track}
\begin{tabular}{cc|c|c|c|c|c}
\hline\hline
\multicolumn{2}{c|}{}                          & MOTA$\uparrow$  & MOTP$\uparrow$  & \ IDS$\downarrow$\  & FRAG$\downarrow$& FN$\downarrow$ \\ \hline
\multicolumn{1}{c|}{\multirow{2}{*}{Single view}} & Day & 0.7382 & 0.5461 & 0   & 47 & 576  \\ \cline{2-7} 
\multicolumn{1}{c|}{}                             & Night & 0.8359 & 0.4531 & 3   & 60 & 395  \\ \hline
\multicolumn{1}{c|}{\multirow{2}{*}{Four views}}  & Day & 0.9255 & 0.6094 & 0   & 21 & 164 \\ \cline{2-7} 
\multicolumn{1}{c|}{}                             & Night& 0.8962 & 0.6217 & 0   & 25 & 253 \\ \hline
\end{tabular}
\end{minipage}\quad
\begin{minipage}{.25\textwidth}
    \centering
    \includegraphics[width=.5\linewidth]{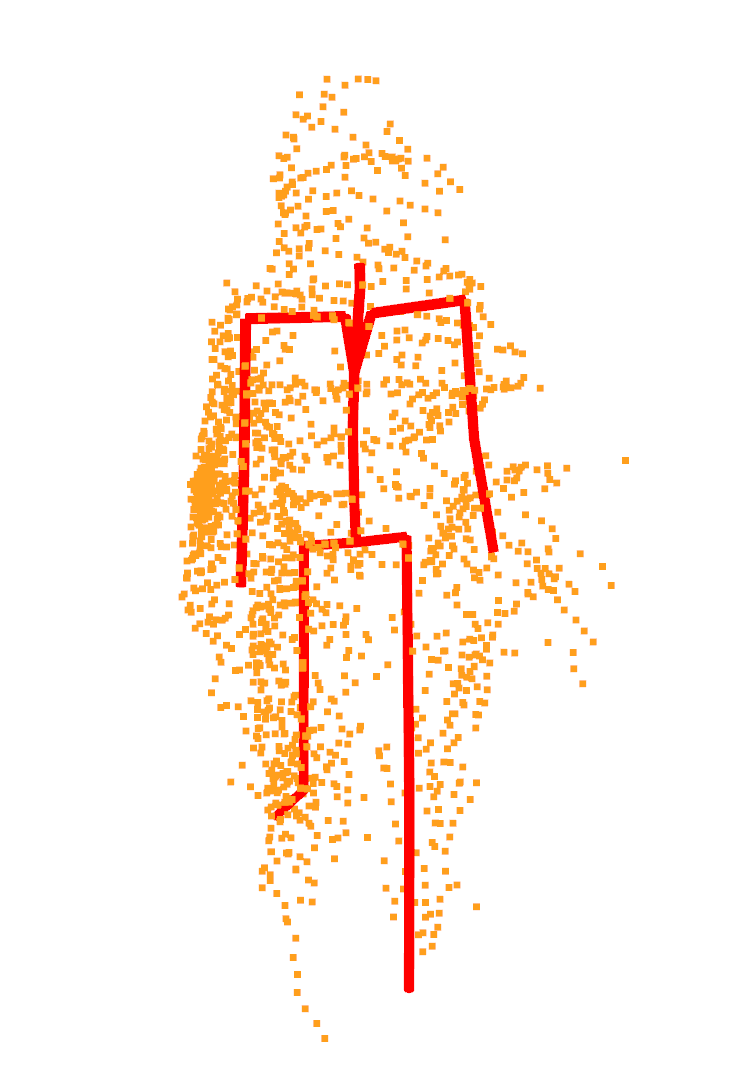}
    \caption{Point clouds with the 3D pose label of a person}\label{fig:3d_pose}
\end{minipage}
\end{figure*}

\begin{figure}
    \centering
    \includegraphics[width=.32\linewidth]{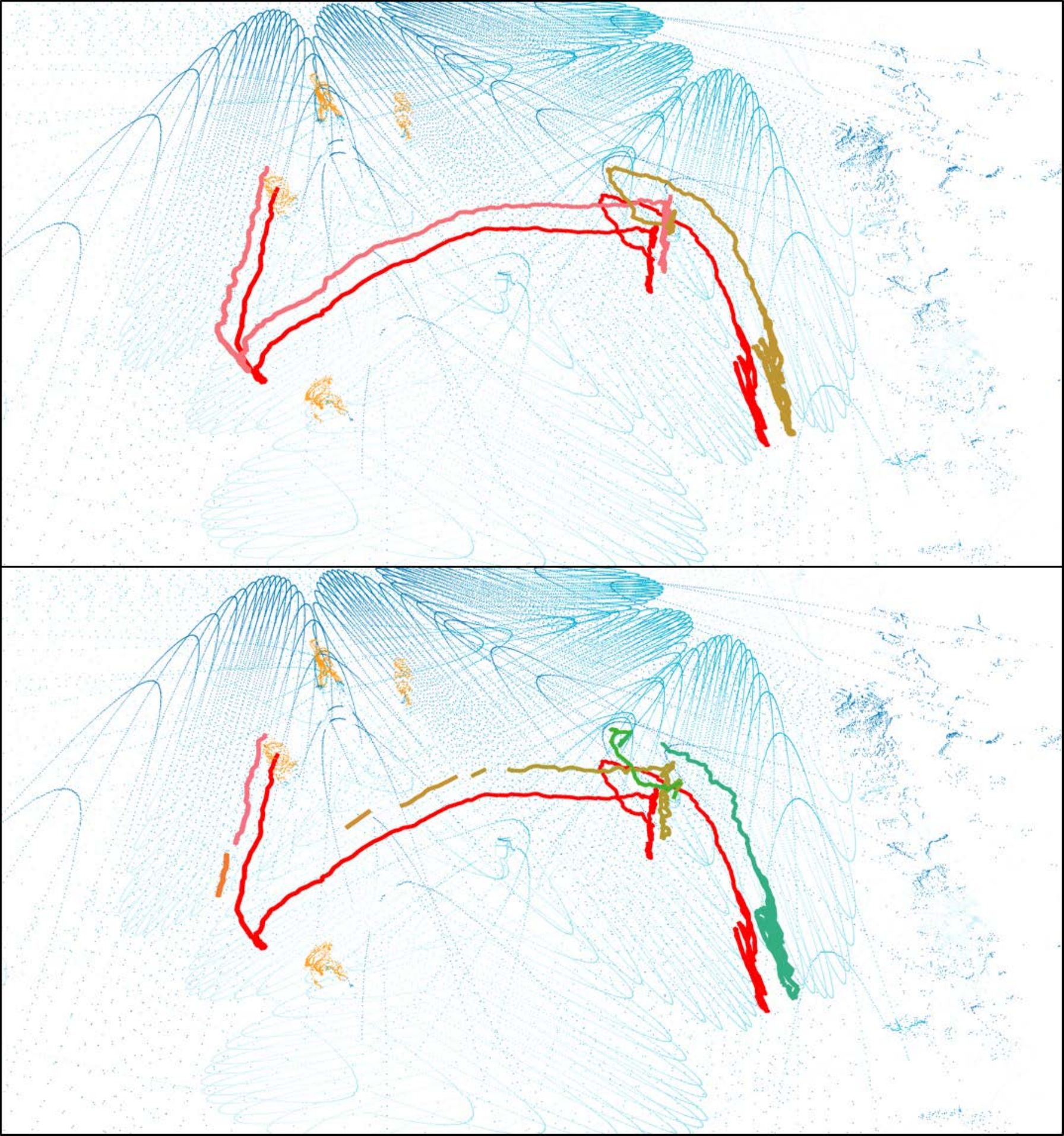}
    \includegraphics[width=.32\linewidth]{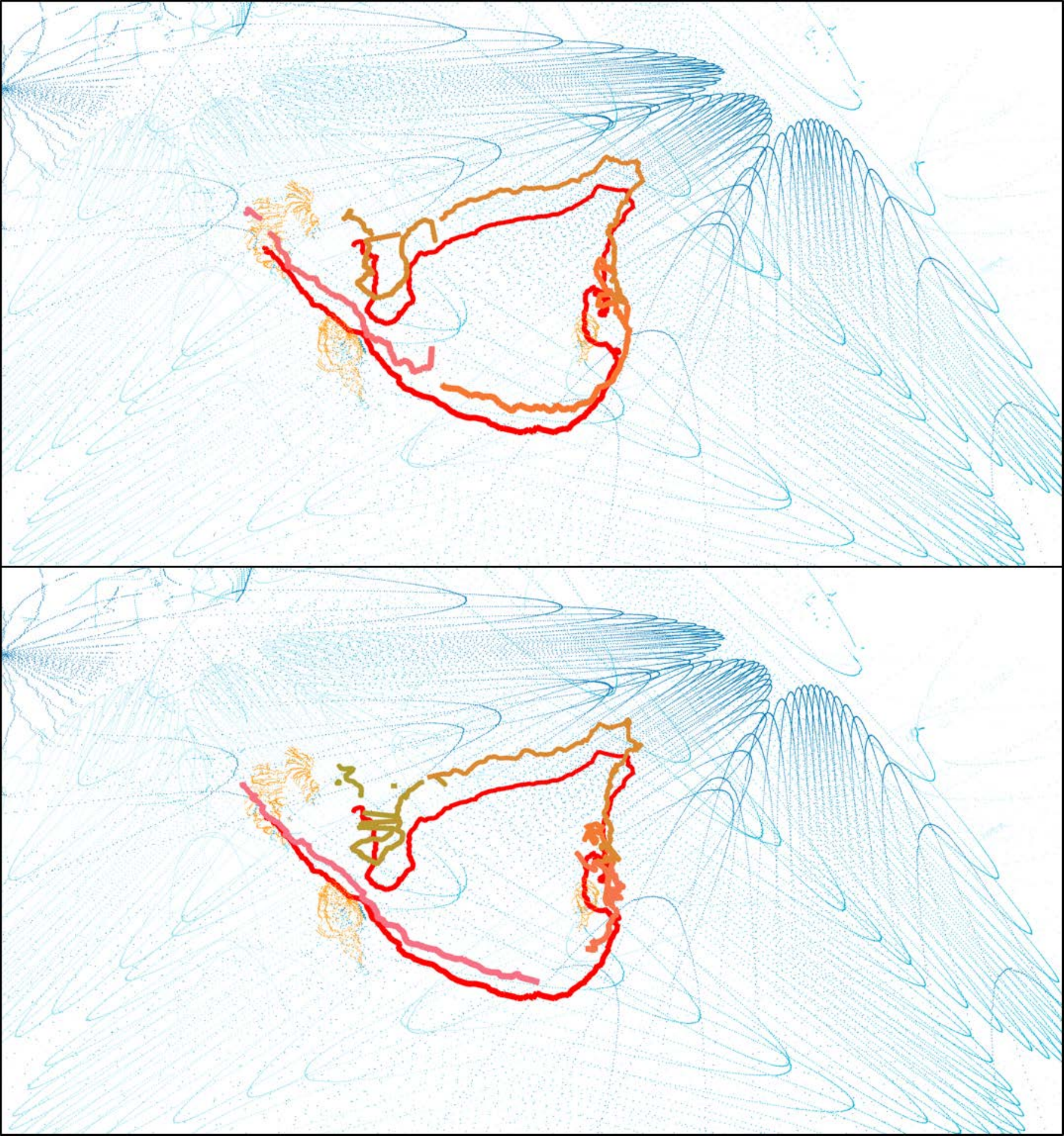}
    \includegraphics[width=.32\linewidth]{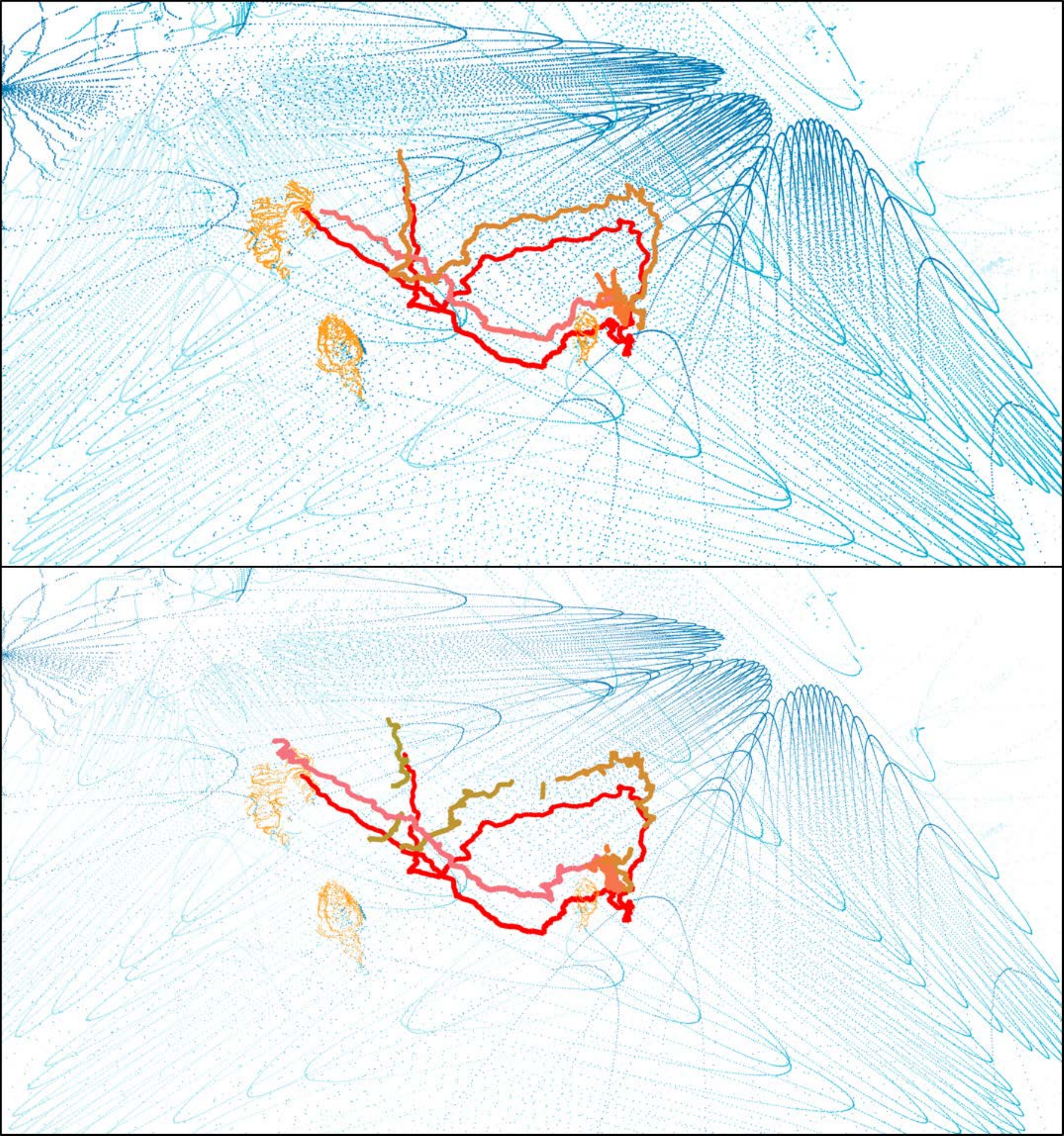}
    \caption{Qualitative tracking results of a single view (bottom) and all four views (top). The ground-truth trajectories are shown by red lines, while the tracking result is represented by others. A change of the line color indicates a fragmentation of the trajectory.}
    \label{fig:track}
    
\vspace{-.3cm}
\end{figure}

As shown in Tab.~\ref{tab:track}, using all four views obtained a higher tracking accuracy and less fragmentation than using a single view. The mainstream 3D multi-target tracking algorithms represented by AB3DMOT~\cite{DBLP:conf/iros/ab3dmot} depend on the detection results. As a result, when using multi-view point cloud, better detection results with a greater recall lead to better tracking results. It can be seen that the long-term tracking of targets with multi-view point clouds is more continuous and unbroken because the effect of occlusion is lessened. False negatives appear less due to the robustness of multi-view point clouds. The ground-truth and tracking results are shown in Fig.~\ref{fig:track}, which shows tracking by multi-view point clouds results in high continuity and accuracy.

In addition, multiple object tracking results based on multi-view point clouds in the day and at night are comparable. Our system employed multiple LiDAR sensors, which are independent of the weather and lighting levels owing to their robustness. When the lighting conditions are very poor at night as shown in Fig~\ref{fig:night}, it is difficult for the RGB cameras to clearly show the objects, but using the point cloud sequence data can still get strong tracking results. By the combination of LiDARs and cameras, our imaging system can be applied in various outdoor scenes.
\section{Limitations}

There is still much room for improvement in our present work. First, the variety and quantity of scenes we tested were not sufficient. More diverse scenes should be captured to test if the spatial calibration technique is reliable. Second, our experiments have focused on point clouds and have not yet utilized RGB images.
Third, since GPS signal can be weak near tall buildings, additional time synchronization techniques need to be developed. Fourth, the experiments were based on existing 3D object detection and tracking algorithms, which are not optimized for the characteristics of our datasets.
Moreover, we have only performed 3D object detection and tracking experiments, although other visual analysis tasks are also possible. For example, the density of human point clouds in our dataset is much higher and more uniform than existing outdoor point cloud datasets, as shown in Fig.~\ref{fig:3d_pose}. It will be an interesting topic to explore point cloud based 3D human pose estimation in large-scale outdoor scenes.

\vspace{-.1cm}
\section{Conclusion}
In this paper, we propose a portable wireless multi-view multi-modal imaging system that is widely applicable to outdoor environments, consisting of a master node and several slave nodes. We utilized multiple LiDARs and cameras to capture multi-view point clouds combined with RGB features. The system is designed to be freely moved and set up for varied scenes. To reduce the labor cost, we propose an automatic spatio-temporal calibration method. Using this 3D imaging system, we collected several multi-view multi-modal datasets for 3D object detection and tracking. We found that using more views improves 3D detection and tracking performance and our system is applicable to different large-scale outdoor scenes. All of the hardware design, codes and datasets have been open sourced.

\vspace{-.3cm}
\subsubsection*{Acknowledgments.} 
The work was supported by the National Key Research and Development Program of China under Grant 2018AAA0102803.

{\small
\bibliographystyle{ieee_fullname}
\bibliography{egbib}
}

\end{document}